\documentclass[pdflatex,sn-mathphys-num]{sn-jnl}

\usepackage{graphicx}
\usepackage{multirow}
\usepackage{makecell}
\usepackage{amsmath,amssymb,amsfonts}
\usepackage{amsthm}
\usepackage{mathrsfs}
\usepackage[title]{appendix}
\usepackage{xcolor}
\usepackage{textcomp}
\usepackage{manyfoot}
\usepackage{booktabs}
\usepackage{array}
\usepackage{listings}
\lstdefinelanguage{PDDL}
{
  sensitive=false,    
  morecomment=[l]{;}, 
  alsoletter={:,-},   
  morekeywords={
    define,domain,problem,not,and,or,when,forall,exists,either,
    :domain,:requirements,:types,:objects,:constants,
    :predicates,:action,:parameters,:precondition,:effect,
    :fluents,:primary-effect,:side-effect,:init,:goal,
    :strips,:adl,:equality,:typing,:conditional-effects,
    :negative-preconditions,:disjunctive-preconditions,
    :existential-preconditions,:universal-preconditions,:quantified-preconditions,
    :functions,assign,increase,decrease,scale-up,scale-down,
    :metric,minimize,maximize,
    :durative-actions,:duration-inequalities,:continuous-effects,
    :durative-action,:duration,:condition, at start, at end, over all
  }
}
\usepackage{pgfplots}
\usepackage{pgfplotstable}
\pgfplotsset{compat=1.18}
\usepackage{tikz}
\usetikzlibrary{patterns}

\usepackage{algorithm}
\usepackage{algpseudocode}

\usepackage{listings}

\usepackage{xcolor}
\usepackage{colortbl}
\usepackage{graphicx}
\usepackage{array}

\usepackage{tikz}
\usetikzlibrary{shapes.geometric, arrows.meta, positioning, fit, backgrounds, calc}

\usepackage{threeparttable}
\usepackage{threeparttablex}   

\newtheorem{definition}{Definition}%
\newcounter{myexample}[section]

\def\BibTeX{{\rm B\kern-.05em{\sc i\kern-.025em b}\kern-.08em
    T\kern-.1667em\lower.7ex\hbox{E}\kern-.125emX}}

\theoremstyle{thmstyleone}%
\theoremstyle{thmstyletwo}%

\theoremstyle{thmstylethree}%

\begin{document}

\title[Article Title]{A Temporal Planning Framework for Disruption Aware Dynamic Route Optimization in Heterogeneous Railway Systems}


\author*[1]{\fnm{Pollob Chandra} \sur{Ray}}\email{22204013@student.duet.ac.bd}

\author[1]{\fnm{Sabah Binte} \sur{Noor}}\email{sabah@duet.ac.bd}

\author[1]{\fnm{
Fazlul Hasan Siddiqui}} \email{siddiqui@duet.ac.bd}

\affil[1]{\orgdiv{Department of Computer Science and Engineering}, 
\orgname{Dhaka University of Engineering \& Technology}, 
\orgaddress{
\city{Gazipur}, 
\postcode{1707},  
\country{Bangladesh}}}


\abstract{
Efficient route optimization plays a vital role in ensuring both safety and punctuality in railway operations.
The challenge is particularly pronounced in heterogeneous multi-gauge railway networks, where varying train speeds, stopping patterns, and infrastructure compatibility constraints increase coordination complexity. 
In single-track systems, these challenges are further intensified because trains share the same track and often require frequent track switching. 
Stochastic disruptions, such as blocked tracks, blocked trains, engine failures, and speed slowdowns, can also introduce additional uncertainty into railway operations and disrupt the timetable.
However, existing studies predominantly focus on high-level timetabling, and often omit operational details such as track switching coordination. Therefore,
this study proposes a framework based on temporal planning for  route optimization and disruption management in heterogeneous railway systems.
The proposed framework formulates railway operations as a temporal planning problem, explicitly modeling gauge compatibility constraints and diverse disruption scenarios.
It generates conflict-free, timestamped operational plans that provide coordinated schedules together with executable action sequences.
To evaluate the proposed framework, we developed a benchmark  consisting of 200 instances with up to 1,000 track points and 120 trains. The framework and the benchmark were evaluated
using two state-of-the-art temporal planners and a plan validator. Our experiments show that the proposed model can generate valid temporal plans across a wide range of heterogeneous railway scenarios with multi-gauge constraints and disruption recovery requirements. These findings demonstrate the feasibility of applying temporal planning techniques to disruption-aware railway routing and scheduling problems.
}


\maketitle

\section{INTRODUCTION}
\label{sec:introduction}
Railway is one of the most efficient ways to transport people and goods at scale, providing  greater energy efficiency and the ability to carry large volumes of freight at lower cost~\cite{islam2024optimizing}.
With rising demand, railway infrastructure encounters significant challenges to ensure reliable and timely services due to network congestion~\cite{bryan2007rail}. 
Thus, maintaining reliable service requires dynamic route optimization and optimal timetable to manage road congestion and reduce operational delays.

Railway scheduling determines train timetables with departure and arrival time for each railway points~\cite{kang2024critical}, while route optimization finds traveling path for trains to minimize travel duration and avoid conflicts~\cite{wang2022train}.
Both have decades of research behind them, genuinely hard to solve at scale and known to NP-hard problem~\cite{leutwiler2023review}.
Real-world systems introduce additional layers of complexity.
For instance, heterogeneous railway systems, in which trains have varying speed profiles and stopping patterns, require a higher level of coordination than uniform metro systems~\cite{yan2019multi}.
Multi-gauge networks amplified this complexity by introducing multiple gauge types for track and train, and require matching gauge types for train movement~\cite{sanchis2021experimental}.
While standard gauge (1435 mm) predominates in Europe and many other regions, several countries operate multi-gauge systems~\cite{sanchis2021experimental, su17188336}. 
For example, Spain railway operates three track gauges: Iberian (1,668 mm) across most of its legacy network, standard (1,435 mm) on high-speed lines, and dual-gauge sections that accommodate both~\cite{sanchis2021experimental}. 
Similarly Bangladesh railway operates three track gauges: meter gauge (1,000 mm), broad gauge (1,676 mm), and dual gauge by combining both~\cite{br2023}.
The practical consequence is that gauge type determines which tracks a train can physically use. 
To drive a train on a specific track must follow the gauge compatibility constraints.
A standard gauge train can only run on standard or dual gauge track. 
An Iberian gauge train needs Iberian or dual gauge. 
In Bangladesh, the same logic holds: meter gauge trains are confined to meter or dual gauge lines, broad gauge to broad or dual gauge.
These constraints significantly limit route options,  thereby increasing operational complexity.

Collectively, these characteristics define what we refer to as a \emph{heterogeneous} railway system. 
We characterize such a system along four dimensions: (i) \emph{speed heterogeneity}, where trains with different speed profiles share the same track segments, (ii) \emph{stopping-pattern heterogeneity}, where each express train may follow a distinct stopping pattern and different passenger boarding times at stations, (iii) \emph{route-choice flexibility}, where multiple admissible paths exist between an origin and a destination, and (iv) \emph{gauge compatibility}, where track segments and rolling stock are constrained by differing gauges.
Most prior studies address only a subset of these dimensions.
In contrast, this work addresses all four jointly.

Moreover, railway operations get disrupted constantly by stochastic events: blocked tracks, blocked trains, engine failures, speed slowdowns~\cite{xiu2024passenger}.
These events deviates the railway schedule (timetable) and require dynamic rescheduling and route optimization to resume its operations.
Numerous studies have been conducted on various aspect of railway operations (discussed in Section~\ref{sec:related_works}), such as scheduling, disruption handling, and rail station management.
However, most studies focused only on selected aspect.
Consequently, most of them generate railway schedule without executable actions, therefore leaving critical decisions like turnout (track switching) to human operators and contribute to safety incidents~\cite{neves2024human}. 
For instance, human errors accounted for 50.77\% of operational incidents in Bangladesh railway during 2022-2023~\cite{br2023}. 
To bridge these gaps, this study proposes a framework based on temporal planning for railway route optimization in heterogeneous systems.

Automated planning is a fundamental field of artificial intelligence that automatically synthesize sequences of actions for achieving specified goals from a given initial state. Temporal planning extends this capability to produce timestamped plans by incorporating action durations and temporal constraints, and minimizes the overall operational time. ~\cite{benton2012temporal}. These timestamps specify precisely which actions to execute and when, making temporal planning particularly well-suited for railway scheduling and route optimization, where precise timing, coordination, and sequencing of train movements are critical. Existing temporal planners can handle the complexity inherent in railway networks efficiently since they employ optimal search algorithms (e.g., A*, IDA*) and domain-independent heuristics. In this work, we encode safety constraints, gauge compatibility, and turnout operations within a temporal planning domain that automates the generation of low-level commands necessary to resolve disruptions. We employ PDDL 2.1 (Planning Domain Definition Language) to design our model.

In this work, we also systematically design a 200-instance benchmark (100 nominal and 100 disrupted instances). The benchmark is generated employing a monotonic scaling across four categories (small, medium, large, and very large), incorporating up to 120 trains and 100 junctions. This enables this study to analyze how the existing planners perform as the network density and disruption severity of the problem increase simultaneously.

We evaluate our proposed framework and benchmark using two state-of-the-art temporal planners: POPF \cite{coles2011popf2}, and OPTIC \cite{benton2012temporal}. 
Experimental results demonstrate that both planners efficiently generate conflict-free plans, with statistical analysis confirming predictable scaling even in highly disrupted scenarios. 
Every generated plan is validated using a temporal validator VAL~\cite{VAL_1374201}, to confirm that the action sequence is consistent and correctly ordered, yielding executable plans ready for real-world deployment.
By providing a formally verifiable planning domain, the proposed framework offers a scalable solution to enhance the safety and efficiency of railway system.

The scope of the proposed framework also has important practical and research implications.
From a practical perspective, the framework targets operational dispatching. Rather than producing only a revised timetable, it generates executable operational plans composed of low-level actions (i.e. train movements, turnout switches, passenger boarding, and recovery operations) with explicit start times and durations, ready for dispatcher-level execution. 
The generated plans minimize the operational time and contain the operational impact of disruptions.  Across our benchmark suite, recovery is achieved with a modest overhead of 6-13\% additional actions under large-scale disruptions. 
Furthermore, disruption management is likewise modeled at the operational level rather than through time-table rescheduling.  An engine failure, for instance, is resolved through an explicit auxiliary-engine rescue pipeline (dispatch, attachment, and assisted movement), whose actions can be executed as issued, instead of leaving the physical recovery process outside the planning model.
This operational feasibility comes at negligible planning cost on commodity hardware, consistent with dispatching horizons of tens of minutes. 
For future research, these boundaries identify where effort is best invested; these findings and directions are examined in detail in Section~\ref{sec:conclusion}.

\subsection{An Illustrative Example}
\begin{figure}[!tbp]
\centering
\begin{tikzpicture}[
    scale=0.62,
    node/.style={
        circle, draw=black!70, very thick,
        minimum size=8mm, font=\small\bfseries, fill=white
    },
    junction/.style={
        diamond, draw=black!70, very thick,
        minimum size=9mm, font=\small\bfseries,
        inner sep=1pt, fill=white
    },
]

\tikzset{
  broad rail/.style  ={line width=1.4pt, black},
  meter rail/.style  ={line width=1.0pt, black!65},
  meter inact/.style ={line width=1.0pt, black!30,
                       dash pattern=on 5pt off 3pt},
  blade/.style       ={line width=1.4pt, black!55, line cap=round},
}

\draw[broad rail] (0, 1.25+0.12) -- (4, 1.25+0.12);
\draw[broad rail] (0, 1.25-0.12) -- (4, 1.25-0.12);
\draw[meter rail] (0, 1.25+0.04) -- (4, 1.25+0.04);

\draw[broad rail] (12, 1.25+0.12) -- (16, 1.25+0.12);
\draw[broad rail] (12, 1.25-0.12) -- (16, 1.25-0.12);
\draw[meter rail] (12, 1.25+0.04) -- (16, 1.25+0.04);

\draw[broad rail] (4+0.036, 1.25+0.115) -- (8+0.036, 0+0.115);
\draw[broad rail] (4-0.036, 1.25-0.115) -- (8-0.036, 0-0.115);

\draw[broad rail] (8+0.036, 0+0.115) -- (12+0.036, 1.25+0.115);
\draw[broad rail] (8-0.036, 0-0.115) -- (12-0.036, 1.25-0.115);

\draw[blade]
    (4, 1.25+0.12) .. controls (4.4,1.25+0.12) and (4.7,1.25+0.00)
    .. (5.0, 1.25-0.10);
\draw[blade]
    (4, 1.25-0.12) .. controls (4.4,1.25-0.12) and (4.7,1.25-0.22)
    .. (5.0, 1.25-0.32);

\draw[blade]
    (11.0, 1.25-0.10) .. controls (11.3,1.25+0.00) and (11.6,1.25+0.12)
    .. (12, 1.25+0.12);
\draw[blade]
    (11.0, 1.25-0.32) .. controls (11.3,1.25-0.22) and (11.6,1.25-0.12)
    .. (12, 1.25-0.12);

\draw[meter inact]
    (4, 1.25+0.04) .. controls (5.0,1.25+0.04) and (6.5,2.0)
    .. (8, 2.5+0.04);
\draw[meter inact]
    (4, 1.25-0.04) .. controls (5.0,1.25-0.04) and (6.5,1.9)
    .. (8, 2.5-0.04);

\draw[meter inact]
    (8, 2.5+0.04) .. controls (9.5,2.0) and (11.0,1.25+0.04)
    .. (12, 1.25+0.04);
\draw[meter inact]
    (8, 2.5-0.04) .. controls (9.5,1.9) and (11.0,1.25-0.04)
    .. (12, 1.25-0.04);

\node[font=\small] at (2,    1.62)  {60 km};
\node[font=\small] at (5.6, -0.45)  {10 km};
\node[font=\small] at (10.4,-0.45)  {15 km};
\node[font=\small] at (5.6,  3.10)  {10 km};
\node[font=\small] at (10.4, 3.10)  {15 km};
\node[font=\small] at (14,   1.62)  {40 km};

\node[node]     (A) at (0,  1.25) {A};
\node[junction] (B) at (4,  1.25) {B};
\node[node]     (C) at (8,  2.5)  {C};
\node[node]     (D) at (8,  0)    {D};
\node[junction] (E) at (12, 1.25) {E};
\node[node]     (F) at (16, 1.25) {F};

\node[draw=black!60, dotted, very thick,
      fit=(C)(D), inner sep=3mm,
      label={[font=\small\itshape]above:Station $s_1$}] {};

\node[anchor=south] (T1img) at (-0.05, 1.5)
    {\includegraphics[width=1.0cm]{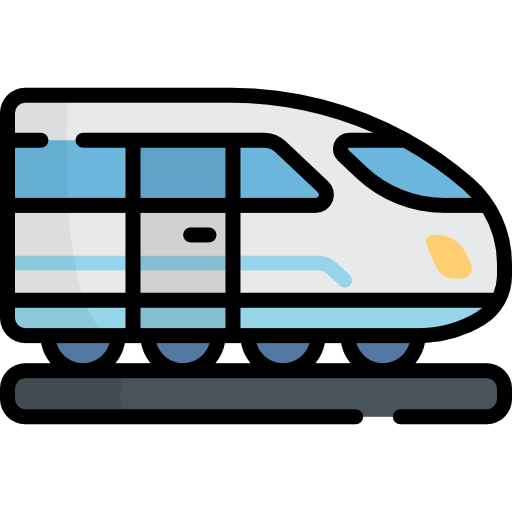}};
\node[anchor=south, font=\small\bfseries] at (T1img.north) {$t_1$};

\node[anchor=south] (T2img) at (16.05, 1.5)
    {\includegraphics[width=1.0cm]{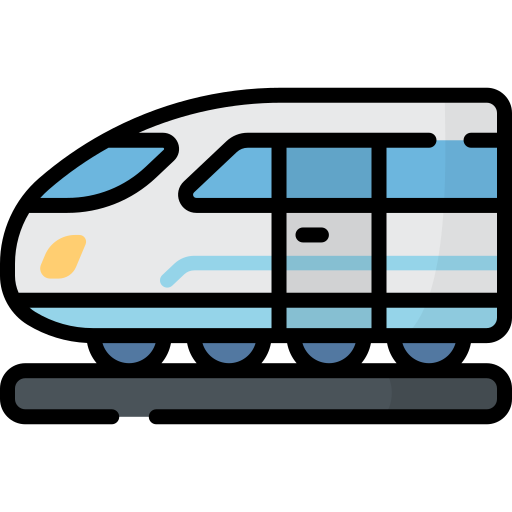}};
\node[anchor=south, font=\small\bfseries] at (T2img.north) {$t_2$};

\begin{scope}[shift={(0,-1.6)}]

  \draw[broad rail] (0,  0.12) -- (1.6,  0.12);
  \draw[broad rail] (0, -0.12) -- (1.6, -0.12);
  \node[font=\footnotesize, right] at (1.75, 0) {Broad gauge};

  \draw[broad rail] (8.6,  0.12) -- (10.2,  0.12);
  \draw[broad rail] (8.6, -0.12) -- (10.2, -0.12);
  \draw[meter rail] (8.6,  0.04) -- (10.2,  0.04);
  \node[font=\footnotesize, right] at (10.35, 0) {Dual gauge};

  \draw[meter rail] (0,  0.04-0.9) -- (1.6,  0.04-0.9);
  \draw[meter rail] (0, -0.04-0.9) -- (1.6, -0.04-0.9);
  \node[font=\footnotesize, right] at (1.75, -0.9) {Meter gauge};

  \draw[meter inact] (8.6,  0.04-0.9) -- (10.2,  0.04-0.9);
  \draw[meter inact] (8.6, -0.04-0.9) -- (10.2, -0.04-0.9);
  \node[font=\footnotesize, right] at (10.35, -0.9)
      {Inactive turnout branch};

  \node[node, minimum size=5mm, font=\tiny\bfseries] at (0.8, -1.8) {P};
  \node[font=\footnotesize, right] at (1.75, -1.8) {Track point};

  \node[junction, minimum size=5mm, font=\tiny\bfseries]
      at (9.4, -1.8) {J};
  \node[font=\footnotesize, right] at (10.35, -1.8)
      {Junction (turnout) point};

  \draw[blade] (0, -2.6) .. controls (0.5,-2.6) and (0.9,-2.75)
      .. (1.3,-2.85);
  \node[font=\footnotesize, right] at (1.75, -2.7)
      {Switch blade (setting)};

  \node[draw=black!60, dotted, very thick, minimum width=8mm,
      minimum height=4.5mm] at (9.4, -2.7) {};
  \node[font=\footnotesize, right] at (10.35, -2.7)
      {Station area};

  \node[anchor=center] at (0.8, -3.55)
      {\includegraphics[width=0.7cm]{fig/train.png}};
  \node[font=\footnotesize, right] at (1.75, -3.6)
      {Train (initial position, heading)};

\end{scope}

\end{tikzpicture}
\caption{Railway network with six track points (A-F). 
Circles denote track points, diamonds denote the junction (turnout) points B and E, with the curved switch blades showing the current turnout setting toward D, and dashed lines indicating the inactive turnout branches. 
Line styles distinguish the gauge of each segment: heavy line pairs denote broad-gauge tracks, lighter line pairs denote meter-gauge tracks, and combined line segments denote dual-gauge tracks (rendered here with three lines for illustration).
The dotted rectangle marks station $s_1$, comprising the platform points C and D. Track distances are labeled along each segment. Train $t_1$ (broad gauge, 60\,km/h) departs from A and train $t_2$ (meter gauge, 66\,km/h) from F, with icons indicating initial positions and headings.}
\label{fig:railway_network1}
\end{figure}

We demonstrate how our framework models railway operations and generates executable plans using a simplified railway network, presented in  Figure~\ref{fig:railway_network1}. The network consists of six track points (\textit{A, B, C, D, E, F}), where \textit{B} and \textit{E} are junction (turnout) points. Junction \textit{B} can connect either to point \textit{C} or \textit{D} (currently connected to \textit{D}), while junction \textit{E} can connect to \textit{C} or \textit{D} (currently connected to \textit{D}).

The network also includes a heterogeneous \textit{multi-gauge} infrastructure. 
Track segments \textit{A-B} and \textit{E-F} are dual gauge, allowing both meter-gauge and broad-gauge trains. 
Segments \textit{B-D} and \textit{D-E} support only broad-gauge trains, while segments \textit{B-C} and \textit{C-E} support only meter-gauge trains. 
Station \textit{S1}, located between junctions \textit{B} and \textit{E}, consists of two parallel platform points \textit{C} and \textit{D} where trains may stop for passenger boarding. 
The network has two trains operating : train \textit{T1} (broad gauge) and train \textit{T2} (meter gauge). 

The task is to move train \textit{t1}  from point \textit{A} to \textit{F},  and to move train \textit{t2}  from point \textit{F} to \textit{A}.  
The train \textit{t1} also needs to stop at point \textit{D} for a 2-minute boarding operation, while train \textit{t2} requires a 3-minute boarding at point \textit{C}.  
Assume, trains \textit{t1} and \textit{t2} travel at 1 km/min (60 km/h) and 1.1 km/min (66 km/h), respectively. Turnout operations require 1 minute to switch track alignment.

\begin{table}[!htbp]
\caption{A temporal plan for the railway routing problem of
Figure~\ref{fig:railway_network1}. Start times and durations are given in minutes and seconds (mm:ss), the planner's $\varepsilon$-separations (0.001\,min), which enforce a strict ordering of simultaneous events, are absorbed by rounding to the nearest second. The plan makespan is 134:39.}
\centering
\renewcommand{\arraystretch}{1.3}
\begin{tabular}{r l r}
\toprule
\textbf{Start (mm:ss)} & \textbf{Action} & \textbf{Duration (mm:ss)} \\
\midrule
00:00 & $\mathsf{DriveTrain}(t_1, A, B, \mathit{broad})$        & 60:00 \\
00:00 & $\mathsf{DriveTrain}(t_2, F, E, \mathit{meter})$        & 36:22 \\
00:00 & $\mathsf{Turnout}(E, D, C)$                             & 01:00 \\
36:22 & $\mathsf{DriveTrain}(t_2, E, C, \mathit{meter})$        & 13:38 \\
50:00 & $\mathsf{Turnout}(E, C, D)$                             & 01:00 \\
50:00 & $\mathsf{BoardPassengers}(t_2, s_1, C)$                 & 03:00 \\
60:00 & $\mathsf{DriveTrain}(t_1, B, D, \mathit{broad})$        & 10:00 \\
70:00 & $\mathsf{Turnout}(B, D, C)$                             & 01:00 \\
70:00 & $\mathsf{BoardPassengers}(t_1, s_1, D)$                 & 02:00 \\
71:00 & $\mathsf{DriveTrain}(t_2, C, B, \mathit{meter})$        & 09:05 \\
72:00 & $\mathsf{DriveTrain}(t_1, D, E, \mathit{broad})$        & 15:00 \\
80:06 & $\mathsf{DriveTrain}(t_2, B, A, \mathit{meter})$        & 54:33 \\
87:00 & $\mathsf{DriveTrain}(t_1, E, F, \mathit{broad})$        & 40:00 \\
\bottomrule
\end{tabular}
\label{tab:plan1}
\end{table}

Table~\ref{tab:plan1} shows the temporal operational plan generated for this railway network problem: a sequence of timestamped actions covering train movements, turnout operations, and passenger boarding, each specified by its start time and execution duration (mm:ss). 
For example, $\mathsf{DriveTrain}(t_1, A, B, \mathit{broad})$ starts at 00:00 with a duration of 60:00, so train $t_1$ departs point $A$ at time zero and arrives at point $B$ after sixty minutes, the traversal time of the segment, the gauge parameter $\mathit{broad}$ records that $t_1$ traverses $A$-$B$ on a compatible track.
The plan also illustrates the temporal coordination the model provides.
Both trains and the turnout at $E$ act concurrently from time zero (00:00), and the junctions are reconfigured as the plan unfolds: the turnout at $E$ is first switched toward $C$ so that $t_2$ can reach its boarding point, and switched back toward $D$ at 50:00 so that $t_1$ can later pass through $E$ on its route to $F$, each switch occupying the junction for its one-minute duration. 
This simple example demonstrates how the generated plan supplies an executable operational schedule that fixes the exact sequence and timing of actions while respecting gauge compatibility, station operations, and turnout constraints.

\subsection{Contributions}
This paper makes the following contributions, organized into modeling, benchmarking, and empirical evaluation:
\begin{enumerate}
\item \textbf{A temporal planning domain for heterogeneous railways: } 
We formalize dynamic route optimization in multi-gauge networks as a temporal planning problem that jointly captures  
(a) gauge compatibility routing constraint,
(b) turnouts as durative, resource-consuming actions, with switching time and junction unavailability factored into the temporal optimization,
and (c) track segments and points as exclusive temporal resources, yielding collision-free schedules by construction.
\item \textbf{Integrated disruption management: } 
We extend the temporal planning framework to support four classes of railway disruptions:: blocked trains, blocked tracks, engine failures, and slowdowns, via dedicated state predicates and durative recovery actions, including an auxiliary-engine rescue pipeline (dispatch, attachment, assisted movement).
\item \textbf{A benchmark problem set.} 
We systematically construct a benchmark of 200
instances for heterogeneous railway systems: 100 nominal and 100 disrupted.
And instances scaling monotonically from small networks to 1{,}000 track points and 120 trains with proportionally increasing disruption severity, enabling fine-grained evaluation of scalability and robustness.
\item \textbf{Empirical Evaluation:} 
We evaluate the benchmark using two off-the-shelf temporal planners (OPTIC and POPF) that solve 99-100\% of the benchmark instances, and all generated plans are verified with the VAL temporal plan validator. 
Our analysis further reveals a size-dependent runtime crossover between the planners and provides a delay decomposition by disruption type, resilience metrics relating disrupted to nominal operations, and supporting statistical analyses.
\end{enumerate}

This work is an extended version of our conference paper accepted at ICSCA 2026. This version substantially expands the work by introducing additional disruption scenarios, a large-scale benchmark problem set, and extended experimental evaluation.

\subsection{Organization}
The remainder of this paper is organized as follows. In Section \ref{sec:related_works}, we discuss the existing approaches to railway scheduling and optimization in the literature. Section~\ref{sec:theoretical_background} provides the necessary background and semantics on temporal planning, and its modeling using PDDL.  In Section ~\ref{sec:methodology}, we present the proposed framework, including the design of the temporal planning domain and the formulation of railway planning problems. Following that,   we discuss experimental evaluation and analysis in Section ~\ref{sec:ex_results} and conclude with our remarks in Section~\ref{sec:conclusion}.

\section{RELATED WORKS}
\label{sec:related_works}
Researchers have proposed a wide range of optimization models and algorithms for railway scheduling, aiming to minimize delays, reduce waiting times, lower operational costs, and enhance system efficiency and service reliability. 
These approaches can be broadly categorized into Maximum Satisfiability (MaxSAT), Mixed Integer Linear Programming (MILP), metaheuristic algorithms, and automated planning. 
MILP and MaxSAT formulate railway scheduling as mathematical optimization problems that produce optimal or near-optimal conflict-free timetables, whereas metaheuristic methods search efficiently for high-quality scheduling solutions under complex constraints. In contrast, temporal planning directly generates executable, timestamped action sequences that naturally capture action durations and concurrency.
The remainder of this section reviews representative studies from each of these four methodological categories and discusses how they address the heterogeneity dimensions introduced in Section~\ref{sec:introduction}.

Maximum Satisfiability (MaxSAT) formulations have recently gained considerable attention for optimizing scheduling under complex operational constraints. Lemos et al.~\cite{lemos2024iterative} proposed an iterative MaxSAT-based framework for the Swiss Federal Railways (SBB) that minimizes delays and routing costs while handling disruptions such as track blockages and speed reductions through rerouting, waiting strategies, and velocity adjustments. 
Gouveia et al.~\cite{gouveia2025maxsat} also proposed a MaxSAT formulation that generates conflict-free timetables while preserving limited flexibility in route choice, tested on real data from SBB and the Washington Metro.
Although these approaches work well within its scope for generating conflict-free schedule, and the framework of Lemos et al.~\cite{lemos2024iterative} additionally supports rerouting and speed adjustment.
However, they generally assume relatively homogeneous   railway infrastructure and rolling stock.
Thus, they generally model trains with broadly uniform physical characteristics and do not jointly capture stopping-pattern variation together with gauge restrictions across heterogeneous rolling stock. 
Furthermore, their primary focus remains to generate conflict free timetable  rather than generation of an executable operational plan composed of low-level, timestamped actions.

Mixed Integer Linear Programming (MILP) has been the most widely adopted method for railway scheduling and rescheduling. 
Chai et al.~\cite{chai2024branch} built a branch-and-cut framework for urban rail that jointly optimizes train sequencing, station timing, and coupling operations under time-varying passenger demand.
Zhuo et al.~\cite{zhuo2024demand} took a similar demand-driven angle, adapting both schedules and train compositions as passenger loads shift. 
Ji et al.~\cite{ji2024optimization} developed a two-stage stochastic optimization model to address uncertainty in railway maintenance durations.  
Additional MILP and ILP-based methods have been proposed for disruption management scenarios, including track blockages, rolling stock failures, and timetable adjustments~\cite{acuna2011sapi, veelenturf2016railway, fischetti2017using, xiu2024passenger, li2025rolling}. 
While these models provide powerful optimization capabilities and in several cases, achieve global optimality for the problems they target.
Their scope is generally confined to timetable and sequencing over a given infrastructure representation. 
However, heterogeneity dimensions such as multi-gauge compatibility and route-level diversity typically fall outside their problem formulations, and the output is an optimized timetable rather than an executable plan of low-level operational actions. 
Nonetheless, they leave open the network-wide, action-level planning problem under combined speed, stopping-pattern, route, and gauge heterogeneity that we address.

Metaheuristic approaches have shown strong performance on scheduling and rescheduling instances~\cite{7160720, Narayanaswami2011}. 
Genetic Algorithms~\cite{nitisiri2019parallel}, Simulated Annealing~\cite{zhang2022train}, Large Neighborhood Search~\cite{20229604362}, and Ant Colony Optimization~\cite{coviello2023integrated} are frequently employed to navigate the non-linear solution spaces of large-scale networks, with reported successes in reducing passenger waiting times and operating cycles under fluctuating demand.
Representative examples include the Problem Space Search metaheuristic of Albrecht et al.~\cite{ALBRECHT2013703}, which reschedules timetables under scheduled maintenance disruptions while preserving schedule quality, and the hybrid framework of D\"{u}ndar and \c{S}ahin~\cite{DUNDAR20131}, which couples a genetic algorithm for conflict resolution with an artificial neural network emulating dispatcher decisions.
In principle, such algorithms are flexible enough to encode heterogeneous infrastructure and rolling-stock constraints within their objective and feasibility functions.
However, these studies frequently rely on simplified network abstractions (i.e. uniform gauge compatibility and homogeneous speed profiles) in order to keep the search space tractable.
This reflects the particular modeling choices made in those studies, not any fundamental limitation of metaheuristics as a class. 
Consequently, while the paradigm itself remains a viable candidate, existing metaheuristic formulations in the literature do not directly produce feasible, action-level plans for the combined heterogeneity addressed here.

Automated planning offers an alternative that generates explicit action sequences for solving a particular task, and has been applied across several domains relevant to ours.
Nyporko et al.~\cite{NYPORKO2025102415} modeled industrial production and manufacturing with PDDL~2.1.
Which select and schedule activities on the resources that perform them, where each action represents an elementary production step. 
Yang et al.~\cite{Yang2022} applied automated planning to emergency-material logistics, regulating the transport and dispatch of supplies to demand points so as to reduce the losses caused by shortages. 
A temporal planning formulation of the temporally constrained journey problem by Caff et. al.~\cite{6984461}. 
The produced solutions satisfying cost, trajectory, and maximum-travel-time constraints. 
Within the railway domain specifically,  Cardellini et al.~\cite{cardellini2021station} developed numeric planning based framework in PDDL+ for train dispatching trains inside stations. 
Louadah et al.~\cite{louadah2021translating} used temporal planning to schedule maintenance in depots by translating ontological knowledge and SWRL rules into PDDL. 
These works demonstrate that planning can coordinate concurrent activities and optimize localized operations. 
However, they are confined to station or depot level settings and do not extend to network-wide route planning, heterogeneous train constraints, or integrated disruption handling, precisely the scope our framework targets.

Across these advancements, two critical gaps remains.
First, most studies incorporate only a subset of the heterogeneity dimensions defined in Section~\ref{sec:introduction}, while multi-gauge compatibility and joint route choice are rarely modeled together. 
Second, the produced output is a conflict-free timetable, and disruption is handled by reactively adjusting that timetable, rather than by proactively generating detailed, timestamped execution plans of low-level actions. 
This study addresses both gaps by introducing a temporal planning framework that explicitly accounts heterogeneity of railway and that produces precise, actionable recovery plans for complex disruptions.

\section{Preliminaries}
\label{sec:theoretical_background}
This section presents the theoretical foundations of temporal planning that underpin our approach. 
It introduces the key concepts and definitions of temporal planning and the Planning Domain Definition Language (PDDL) required for modeling the proposed framework.

\subsection{Temporal Planning}
Automated planning and scheduling is a core subfield of artificial intelligence concerned with the automatic generation of action sequences that transform a given initial state into one satisfying specified goal conditions \cite{ghallab2004automated}. Learning-based approaches derive decision-making behavior from data through statistical training, whereas automated planning relies on explicit, symbolic models of actions, states, and goals, and computes solutions through search and reasoning over these models. 
A planning system (planner), such as Optic or POPF, takes as input a formal description of the initial state, the available actions, and the goal conditions, and produces a plan (a structured sequence of actions), that achieves the goals when executed.


In classical planning, actions are assumed to be instantaneous and sequentially executed. The environment is typically modeled as fully observable, deterministic, static (i.e., state changes occur only through agent actions), and finite. A solution to a classical planning problem is a sequence of actions (i.e., a plan) such that executing the actions successively from the initial state results in a state that satisfies the goal condition. However, modeling many real-world domains that involve action durations, temporal dependencies, or concurrent activities can be challenging within the classical planning framework.
Railway operations, for example, involve activities such as train movements and turnout operations that require explicit representation of time and resource constraints.

Temporal planning extends classical planning by explicitly incorporating time into the model~\cite{cenamor2019predictability}.
Intuitively, a temporal planning problem (Definition~\ref{def:temporal-problem}) specifies the system state using logical and numeric variables, a set of durative actions that modify the state, an initial configuration of the system (given by an assignment over both fluent types) and a set of goal conditions that the planner must achieve. 
A durative action (Definition~\ref{def:durative-action}) represents an activity that unfolds over a time interval bounded by a start and an end instant.
Three temporally distinguished sets of conditions govern its applicability: conditions required at the start instant, conditions required to hold throughout the entire interval, and conditions required at the end instant.
Correspondingly, its effects are partitioned into those applied at the start instant and those applied at the end instant, reflecting how the action's impact on the state may unfold at different points in its execution.

\begin{definition}\label{def:temporal-problem}
A \textbf{temporal planning problem} is defined as a 5-tuple $P_T = \langle F, V, A_d, I, G \rangle$, where:
\begin{itemize}
    \item $F$ is a finite set of propositional fluents,
    \item $X$ is a set of real-valued numeric fluents,
    \item $A_d$ is a finite set of \emph{durative actions},
    \item  $I \subseteq F \cup X$ represents the initial state, given by a truth assignment to the fluents in $F$ and a real-valued assignment to the numeric fluents in $X$, and 
    \item $G \subseteq F \cup X$ represents the goal condition, expressed as a set of propositional fluents that must hold (optionally together with numeric constraints over $X$)
\end{itemize} 
\end{definition}

\begin{definition}\label{def:durative-action}
A \textbf{durative action} $a \in A_d$ is defined as
$a=\langle \mathit{pre}_{\vdash}(a), \mathit{pre}_{\leftrightarrow}(a), \mathit{pre}_{\dashv}(a), \mathit{eff}_{\vdash}(a), \mathit{eff}_{\dashv}(a), \mathit{dur}(a) \rangle$, where
\begin{itemize}
    \item $\mathit{pre}_{\vdash}(a)$ are \emph{start conditions}, required to hold at the instant the action starts,
    \item $\mathit{pre}_{\leftrightarrow}(a)$ are \emph{over-all conditions}, required to hold continuously throughout the action's execution duration,
    \item $\mathit{pre}_{\dashv}(a)$ are \emph{end conditions}, required to hold at the instant the action ends,
    \item $\mathit{eff}_{\vdash}(a)$ are \emph{start effects (changes)}, applied at the instant the action starts,
    \item $\mathit{eff}_{\dashv}(a)$ are \emph{end effects (changes)}, applied at the instant the action ends, and
    \item $\mathit{dur}(a) \in \mathbb{R}^+$ denotes the action duration.
\end{itemize}
\end{definition}

Unlike classical planning, a temporal plan is not just a sequence of actions but a time-stamped schedule of $(t_i, a_i)$ pairs, where $t_i$ denotes the start time of action $a_i$ \cite{haslum2019introduction}.
A temporal plan is valid if and only if (i) all preconditions are satisfied at their required times, (ii) over all (invariant) conditions hold throughout $[t_i, t_i + \text{dur}(a_i)]$, no conflicting effects occur simultaneously, and the goal $G$ is satisfied at plan completion: $t_{\text{end}} = \max_i(t_i + \text{dur}(a_i))$. The makespan (Definition \ref{def:makespan}) of a temporal plan measures the total time required to complete all actions, from the start of the earliest action to the completion of the latest one.

\begin{definition}\label{def:makespan}
The \textbf{makespan} of a temporal plan $\Pi_T$ is the total execution time of the plan, defined as
\[
\mathrm{makespan}(\Pi_T) 
= \max_{(a_i, t_i) \in \Pi_T} \left( t_i + \mathrm{dur}(a_i) \right).
\]
\end{definition}

To formalize a planning problem and generate valid temporal plans, we need to describe the planning environment, such as states, actions, and goals. The Planning Domain Definition Language (PDDL) \cite{haslum2019introduction} is the most widely adopted language for this purpose. 

\subsection{Planning Domain Definition Language}
\label{sec:pddl}
An Automated planning system requires a formal representation of the \emph{initial state}, the \emph{actions} to enable transitions between states, and the \emph{goal conditions} to generate a valid plan. For the past two decades, the planning community has widely adopted the Planning Domain Definition Language (PDDL) as the standard for modeling planning problems. PDDL divides a planning problem into two components: a domain and a problem instance. The domain captures the general structure of the planning environment, including state predicates and action definitions, while the problem instance specifies a particular scenario by defining the initial state and the desired goal conditions.

PDDL 2.1 extends classical PDDL by introducing temporal and numeric features \cite{fox2003pddl2}. 
In particular, it supports durative actions with explicit durations, temporal qualifiers (\texttt{at start}, \texttt{over all}, \texttt{at end}), numeric fluents and arithmetic expressions, and optimization metrics (e.g., makespan minimization) .

We describe the syntax of PDDL 2.1 using the \emph{domain} definition and the \emph{problem instance} of a toy railway system, presented in Listings \ref{lst:domain} and \ref{lst:problem}, respectively.

\paragraph{Domain Defintion}
A temporal planning domain specifies the object types, state predicates, numeric fluents, and durative-action schemas of the environment. Each durative action declares typed parameters, a duration, temporally annotated conditions that must hold \emph{at start}, \emph{over all} (as invariants), or \emph{at end} of execution, and effects applied at the start or end of the action. Listing~\ref{lst:domain} presents a simplified railway domain illustrating these concepts. The domain defines two object types, \emph{train} and \emph{location}, two predicates, at and connected, representing the train's current location and the connectivity between locations, respectively, a numeric fluent total-cost, and a single durative action, \texttt{move}. The action has a fixed duration of five time units and requires the train to be at the origin location when execution begins. Also, the origin and destination locations need to be connected. At the start of execution, the action removes the train from the origin and places it at the destination upon completion. It then increases the accumulated travel cost by 10 units at the end of the action. Although this example uses a fixed action duration for simplicity, PDDL~2.1 also permits duration expressions over numeric fluents. Our proposed railway domain (Section~\ref{sec:methodology}) computes action durations dynamically from track distances, train speeds, and slowdown factors.

\begin{lstlisting}[
  language=PDDL,
  caption = {Temporal Domain definition for a toy \textit{railway} system},
  basicstyle=\small,
  captionpos=b, label ={lst:domain}]
(define (domain simple_temporal_railway)
    (:requirements :strips :typing :durative-actions :fluents)
    (:types train location)
    (:predicates
        (at ?t - train ?l - location)
         (connected ?from - location ?to - location))
    (:functions
        (total-cost))
    (:durative-action move
    :parameters (?t - train ?from - location ?to - location)
    :duration (= ?duration 5)
    :condition (and
        (at start (at ?t ?from))
        (at start (connected ?from ?to)))
    :effect (and
        (at start (not (at ?t ?from)))
        (at end (at ?t ?to))
        (at end (increase (total-cost) 10)))
)
)
\end{lstlisting}

\paragraph{Problem Definition}
A problem instance declares the objects, the initial state, and the goal of a concrete scenario, providing the data from which the planner instantiates the domain actions. 
Listing~\ref{lst:problem} shows a toy instance for the domain in Listing~\ref{lst:domain}. It defines one train, three locations, the initial location of the train, the connectivity between adjacent locations, and initializes the accumulated cost to zero. The goal specifies that the train must reach location \textit{C}, while the metric directs the planner to minimize the total action cost.
Alternatively, specifying the special fluent \emph{total-time} as the optimization metric minimizes the plan makespan. Temporal planners then search for a temporally valid plan optimizing the given objective.
\begin{lstlisting}[
  language=PDDL,
  caption = {Problem definition for the railway domain in Listing \ref{lst:domain}} ,
  basicstyle=\small,
  captionpos=b, label ={lst:problem}]
(define (problem toy_instance)
    (:domain simple_temporal_railway)
    (:objects
        train1 - train
        A B C - location)
    (:init
        (at train1 A)
        (connected A B)
        (connected B C)
        (= (total-cost) 0))
    (:goal
        (at train1 C))
    (:metric minimize (total-cost))
)
\end{lstlisting}

A temporal plan for this instance consists of two sequential durative actions. Since locations \texttt{A} and \texttt{C} are not directly connected, the train must travel via the intermediate location \texttt{B}. Each plan entry specifies the action's start time, its ground instantiation, and its duration. The temporal planners used in this work produce plans in the following format:
\[
\pi =
\bigl\{
(0.0,\ \mathsf{Move}(\mathit{train1},A,B),\ 5.0),\;
(5.0,\ \mathsf{Move}(\mathit{train1},B,C),\ 5.0)
\bigr\}.
\]
This plan has a makespan of $10$ time units and a total cost of $20$, since each action has a duration of $5$ time units and incurs a cost of $10$ units.

Such expressive capabilities of PDDL~2.1, including durative actions with temporal annotations and numeric fluents, provide a solid basis for the railway domain model developed in this thesis

\section{METHODOLOGICAL FRAMEWORK AND PROBLEM FORMULATION}
\label{sec:methodology}
This section presents the proposed \emph{Disruption Aware Railway Temporal Planning (DART)} framework, which models railway systems as a temporal planning problem. The framework generates detailed, time-stamped plans for coordinating train movements, infrastructure operations, and disruption recovery while satisfying gauge compatibility and operational constraints.

To address operational challenges in heterogeneous railway networks, DART incorporates a disruption taxonomy and corresponding recovery strategies for each disruption type. The railway network, rolling stock, infrastructure components, and disruption scenarios are encoded in PDDL~2.1, enabling temporal planners to generate executable plans that dynamically adapt to operational disruptions.

The remainder of this section presents the disruption taxonomy, the formal temporal planning model, the PDDL domain and problem representations, the railway action models, and the benchmark generation methodology.

\subsection{ Disruption Taxonomy and Recovery Strategies}
\label{sec:disruption_senarios}
Railway operations in heterogeneous multi-gauge networks are frequently affected by stochastic disruptions that deviate from nominal schedules, thus requiring recover planning. 
DART formalizes four primary disruption types, each represented by dedicated predicates and durative recovery actions (Section~\ref{sec:domain_construction}). 
These disruptions also form the basis for our benchmark instances in Section~\ref{sec:benchmark_generation}:

\begin{enumerate}
    \item \textbf{Blocked Track Disruption} occurs when a railway track segment becomes unavailable for train movements due to accidents, infrastructure failures, or maintenance activities.  Recovery strategies in our framework include (a) waiting at the current position until the track is cleared, or (b) rerouting via gauge-compatible alternative paths using turnout operations. 
\item \textbf{Blocked Train Disruption}
 occurs when a train is temporarily immobilized (e.g., due to logistical issues, accidents, or crew unavailability). When this disruption occurs, the affected train must remain at the disruption point until the blockage is resolved,while other trains continue operations
\item \textbf{Slowdown Disruption} reduces train speeds over specific track segments due to external conditions such as adverse weather. Travel time on a disrupted segment is computed as:  

\begin{equation}
\label{eq:travelling_time_slowdown}
\text{traveling-time} = \frac{\text{track-distance}(D, E)}{(1 - \alpha) \cdot \text{train-speed}(t)}
\end{equation}
where $t$ denotes the train traversing the disrupted segment, and $\alpha \in (0, 1]$ denotes the slowdown factor (e.g., $\alpha = 0.2$ for a 20\% speed reduction). Recovery strategies include traversing with increased travel time or rerouting to minimize makespan.  
\item \textbf{Engine Failure Disruptions} occurs when a locomotive becomes inoperative. Recovery involves dispatching an auxiliary engine to the affected train and coupling it to resume operations. 
\end{enumerate}
Together, these four disruption categories constitute a comprehensive operational stress model for the heterogeneous railway networks.

\subsection{Temporal Railway Task Formulation}
We formulate railway route optimization and disruption management as a temporal planning problem. The formulation assumes that the railway system is fully observable and deterministic, all trains must reach their designated destinations and satisfy terminal conditions, disruption durations (e.g., blocked tracks, blocked trains, and slowdowns) are known a priori, and trains travel at constant speeds without acceleration or deceleration. For clarity and accessibility, we  present the railway planning model using a first-order logical formalization, allowing readers to understand the underlying model independently of its PDDL implementation. The corresponding PDDL domain and problem specifications are then derived from this formalization.

\begin{definition}
A railway temporal planning task is represented as

\[
P_R = \langle F_R, X_R, A_R, I_R, G_R \rangle
\]

where:

\begin{itemize}
\item $F_R$ is the finite set of grounded propositional facts over the predicate set
$\mathcal{P}_R$,describing the railway state.

\[
\mathcal{P}_R =
\left\{
\begin{array}{l}
\textit{train\_at},\ \textit{engine\_at},\ \textit{connected},\ \textit{free},
\textit{track\_accessible},\ \textit{junction\_point},\\
\textit{turnout\_alternative},\ \textit{train\_gauge},
\textit{track\_gauge},\ \textit{engine\_gauge},\\
\textit{track\_blocked},
\textit{track\_clear},
\textit{train\_blocked},\ \textit{train\_clear},
\textit{engine\_damaged},\\ \textit{engine\_functional},
\textit{engine\_free},\ \textit{engine\_attached},
\textit{platform\_at},\\ \textit{boarding\_point}, 
\textit{passengers\_boarded}
\end{array}
\right\};
\]
\item $X_R$ is the finite set of grounded numeric fluents over the function set
$\mathcal{X}_R$, representing quantitative properties of the railway system.

\[
\mathcal{X}_R =
\left\{
\begin{array}{l}
\textit{distance},
\textit{train\_speed},
\textit{engine\_speed},
\textit{boarding\_time},
\textit{turnout\_time},\\
\textit{slowdown},
\textit{train\_blockage\_time},
\textit{track\_clear\_time},
\textit{engine\_attach\_time}
\end{array}
\right\};
\]
\item $A_R$ is the finite set of durative actions representing railway operations.

\[ A_R=\left \{ 
\begin{array}{l}
\mathsf{DriveTrain}, \mathsf{DriveEngine}, \mathsf{DriveAssistedTrain}, \mathsf{BoardPassengers}, \mathsf{Turnout},\\ \mathsf{ResolveTrainBlockage}, \mathsf{ClearBlockedTrack}, \mathsf{AttachEngine},\\ \mathsf{DriveEngineToDamagedUpTrain}, \mathsf{DriveEngineToDamagedDownTrain}
\end{array}
\right\};
\]

\item $I_R$ denotes the initial railway state; and
\item $G_R$ specifies the goal conditions that every valid plan must satisfy.
\end{itemize}

\end{definition}
The goal of the planning task is to ensure that every train reaches its designated destination while completing all required passenger-service operations. Depending on the disruption scenario, additional recovery conditions, such as clearing blocked tracks or recovering disabled trains, may also be required. Accordingly, the goal condition is defined as

\[
G_R=
\left(
\bigwedge_{t\in T}
train\_at(t,\mathrm{Dest}(t))
\right)
\wedge
\left(
\bigwedge_{t\in T}
passengers\_boarded(t)
\right).
\]

Among all valid temporal plans satisfying the goal conditions, the planner seeks the schedule with the minimum makespan,

\[
\min \; \textit{Makespan}(\pi),
\]

where $\pi$ denotes the generated temporal plan.

\subsubsection{World Representation}
This subsection defines the type hierarchy and state predicates used to model the railway network, the operational entities within it, and the numeric fluents governing action durations.

\paragraph{Type System}
The domain is organized over five object types: \emph{track-point}, \emph{train}, \emph{gauge-type}, \emph{engine}, and \emph{station}. Track points form the nodes of the railway network and, together with the segments connecting them, define its topology. Trains and auxiliary engines represent the rolling stock; each train and engine has an associated gauge type and a cruising speed. Stations group the platform points where passenger boarding occurs. Gauge types capture the physical gauge classifications (meter, broad, dual) that constrain which trains may use which segments.

\paragraph{State Predicates}
The predicates $\mathcal{P}_R$ collectively define the operational world state of the railway environment. They are organized into five groups by the aspect of the railway they describe.

\begin{itemize}
\item \textbf{Spatial configuration.} \textit{train\_at} and \textit{engine\_at} locate trains and auxiliary engines at track points; \textit{platform\_at} and \textit{boarding\_point} identify where a train must stop for service, and \textit{passengers\_boarded} records that service is complete; \textit{junction\_point} marks a point as a turnout.
\item \textbf{Infrastructure connectivity.} \textit{free} marks a point as unoccupied; \textit{connected} reflects the currently active turnout routing between two points; \textit{track\_accessible} marks a segment as available for traversal, and is withdrawn while a vehicle occupies it, making segments exclusive temporal resources; \textit{turnout\_alternatives} records the two routing choices available at a junction.
\item \textbf{Gauge compatibility.} \textit{train\_gauge}, \textit{engine\_gauge}, and \textit{track\_gauge} attach a gauge type to a train, an auxiliary engine, and a track segment, respectively, so that movement actions can enforce compatibility directly in their preconditions.
\item \textbf{Disruption state.} \textit{track\_blocked} and \textit{track\_clear} are complementary facts describing whether a segment is passable; \textit{train\_blocked} marks an immobilized train; \textit{engine\_damaged}, \textit{engine\_free}, and \textit{engine\_attached} track a locomotive failure and its recovery via an auxiliary engine.
\item \textbf{Directional coordination.} \textit{up} and \textit{down} orient a segment; \textit{heading\_up} and \textit{heading\_down} record a train's or engine's current travel direction, so a rescue engine can be dispatched against the disabled train's heading.
\item \textbf{Constants.} A set of numeric fluents $\mathcal{X}_R$ governs action durations and remains fixed for a given instance: segment \textit{distance} and vehicle \textit{train\_speed}/\textit{engine\_speed} determine travel time; \textit{boarding\_time} and \textit{turnout\_time} fix service and switching durations; \textit{slowdown} scales travel time under a weather- or track-condition disruption; and \textit{train\_blockage\_time}, \textit{track\_clear\_time}, and \textit{engine\_attach\_time} fix the duration of the corresponding recovery actions. Each is computed from infrastructure and vehicle properties rather than hardcoded, so a single domain covers heterogeneous networks without per-instance modification.
\end{itemize}
\subsubsection{Modeling Decisions}
The following decisions shape the predicates and action schemas introduced below, balancing expressiveness against planning tractability.

\textbf{Segments and points as exclusive temporal resources.} Rather than checking collisions post hoc, track segments and points are made temporarily unavailable for the duration a vehicle occupies them, so every plan is collision-free by construction rather than by validation.

\textbf{Declarative gauge compatibility.} Gauge type is attached to trains, engines, and track segments as static facts, and every movement action's preconditions require a matching gauge. This keeps gauge reasoning local to each action rather than requiring a separate consistency check over the whole route.

\textbf{Disruptions as explicit state, not stochastic events.} Blocked tracks, blocked trains, engine damage, and slowdowns are represented as ordinary state predicates and a fluent, not sampled during execution. This lets a temporal planner reason about recovery the same way it reasons about nominal operations, using the same search machinery.

\textbf{Directional dispatch for engine rescue.} Auxiliary-engine dispatch is split into an up-bound and a down-bound schema rather than one direction-agnostic action, so that a rescue engine is only ever routed to approach a disabled train against its direction of travel, matching how rescue operations work in practice.
\subsubsection{Durative Actions}
The proposed railway domain is modeled using ten durative actions to represent operational activities and disruption recovery procedures. We categorize  the railway operations into four types: movement operations, passenger service operations, disruption recovery operations, and infrastructure operations. 

\subparagraph{Movement Operations}
The $\mathsf{DriveTrain}$ and $\mathsf{DriveEngine}$ actions (Table~\ref{tab:actions_movement}) model the traversal of trains and auxiliary engines between railway points, both share the same structure.
We describe $\mathsf{DriveTrain}$ in detail as the representative example and highlight only the distinguishing features of subsequent actions. 
Its parameters are a train $t$, a track segment defined by the points $n_i$ and $n_j$, and a gauge type $g$.

The temporal conditions of $\mathsf{DriveTrain}$ ensure that, at initiation, the train and track are gauge-compatible, the segment is clear and accessible, and the train is neither blocked nor operating with a damaged engine. 
The destination point must remain free throughout execution and upon completion, preventing conflicting occupancy. 
The temporal effects mark the segment inaccessible in both directions at the start of the action, avoiding simultaneous use.
Upon completion, the destination is marked occupied and segment accessibility is restored in both directions.

\begin{table*}[htbp]
\centering
\caption{Movement-operation action schemas. Parameter shorthands: $t$ a
train; $e$ an auxiliary engine; $n_i$, $n_j$ track points; $g$ a gauge
type. $\tau_{u}(n_i,n_j) = d(n_i,n_j)/\bigl(u\,(1-\alpha(n_i,n_j))\bigr)$
denotes the traversal time of segment $(n_i,n_j)$ at speed $u$
(Eq.~\eqref{eq:travelling_time_slowdown}).}
\label{tab:actions_movement}
\footnotesize
\setlength{\tabcolsep}{5pt}
\begin{tabular}{l l l l l l}
\toprule
\textbf{Action} & \multicolumn{2}{l}{\textbf{Preconditions}}
                & \multicolumn{2}{l}{\textbf{Effects}}
                & \textbf{Duration} \\
\cmidrule(lr){2-3}\cmidrule(lr){4-5}
 & ann. & predicate & ann. & predicate & \\
\midrule
$\mathsf{DriveTrain}$
 & Start & $\mathit{train\_at}(t,n_i)$
 & Start & $\neg\mathit{train\_at}(t,n_i)$
 & $\tau_{v(t)}(n_i,n_j)$ \\
$(t,n_i,n_j,g)$
 & & $\mathit{train\_gauge}(t,g)$
 & & $\mathit{free}(n_i)$ & \\
 & & $\mathit{track\_gauge}(n_i,n_j,g)$
 & & $\neg\mathit{track\_accessible}(n_i,n_j)$ & \\
 & & $\neg\mathit{train\_blocked}(t,n_i)$
 & & $\neg\mathit{track\_accessible}(n_j,n_i)$ & \\
 & & $\neg\mathit{engine\_damaged}(t,n_i)$ & & & \\
 & & $\mathit{track\_clear}(n_i,n_j)$ & & & \\
 & & $\mathit{track\_accessible}(n_i,n_j)$ & & & \\
\addlinespace[5pt]
 & Over-all & $\mathit{connected}(n_i,n_j)$
 & End & $\mathit{train\_at}(t,n_j)$ & \\
 & & $\mathit{free}(n_j)$
 & & $\neg\mathit{free}(n_j)$ & \\
\addlinespace[5pt]
 & End & $\mathit{free}(n_j)$
 & & $\mathit{track\_accessible}(n_i,n_j)$ & \\
 & & & & $\mathit{track\_accessible}(n_j,n_i)$ & \\
\midrule
$\mathsf{DriveEngine}$
 & Start & $\mathit{engine\_at}(e,n_i)$
 & Start & $\neg\mathit{engine\_at}(e,n_i)$
 & $\tau_{v_a(e)}(n_i,n_j)$ \\
$(e,n_i,n_j,g)$
 & & $\mathit{engine\_free}(e)$
 & & $\mathit{free}(n_i)$ & \\
 & & $\mathit{engine\_gauge}(e,g)$
 & & $\neg\mathit{track\_accessible}(n_i,n_j)$ & \\
 & & $\mathit{track\_gauge}(n_i,n_j,g)$
 & & $\neg\mathit{track\_accessible}(n_j,n_i)$ & \\
 & & $\mathit{track\_clear}(n_i,n_j)$ & & & \\
 & & $\mathit{track\_accessible}(n_i,n_j)$ & & & \\
\addlinespace[5pt]
 & Over-all & $\mathit{connected}(n_i,n_j)$
 & End & $\mathit{engine\_at}(e,n_j)$ & \\
 & & $\mathit{free}(n_j)$
 & & $\neg\mathit{free}(n_j)$ & \\
\addlinespace[5pt]
 & End & $\mathit{free}(n_j)$
 & & $\mathit{track\_accessible}(n_i,n_j)$ & \\
 & & & & $\mathit{track\_accessible}(n_j,n_i)$ & \\
\bottomrule
\end{tabular}
\end{table*}

\subparagraph{Passenger Service Operations}
Passenger service activities are represented through the \verb|board-passengers| action (Table~\ref{tab:actions_service}), which ensures that departure occurs only after boarding is complete. 
Boarding time is defined by $\beta(t,s)$, the time (in minutes) required for train $t$ to board passengers at station $s$.
\begin{table}[htbp]
\centering
\caption{Passenger-service action schema. $s$ denotes a station; other
shorthands as in Table~\ref{tab:actions_movement}.}
\label{tab:actions_service}
\footnotesize
\setlength{\tabcolsep}{5pt}
\begin{tabular}{l l l l l l}
\toprule
\textbf{Action} & \multicolumn{2}{l}{\textbf{Preconditions}}
                & \multicolumn{2}{l}{\textbf{Effects}}
                & \textbf{Duration} \\
\cmidrule(lr){2-3}\cmidrule(lr){4-5}
 & ann. & predicate & ann. & predicate & \\
\midrule
$\mathsf{BoardPassengers}$
 & Start & $\mathit{train\_at}(t,n)$
 & End & $\mathit{passengers\_boarded}(t,n)$
 & $\beta(t,s)$ \\
$(t,s,n)$
 & & $\mathit{boarding\_point}(t,n)$ & & & \\
 & & $\mathit{platform\_at}(n,s)$ & & & \\
\addlinespace[5pt]
 & Over-all & $\mathit{train\_at}(t,n)$ & & & \\
\bottomrule
\end{tabular}
\end{table}

\subparagraph{Disruption Recovery Operations}
 To support disruption recovery, \verb|drive-engine-to-damaged-up-train| and \verb|drive-engine-to-damaged-down-train| actions dispatch auxiliary engines toward immobilized trains while respecting directional and connectivity constraints. These actions are similar in structure to movement operations. 
 
The \verb|attach-engine| action (Table \ref{tab:actions_recovery}) models the coupling of an auxiliary engine to a damaged train on a track segment. Both entities (the corresponding engine and train) must be co-located throughout the operation, enforced via \verb|over| all conditions. The engine is marked unavailable at the start \verb|(not (engine-free ?en))|, and upon completion \verb|engine-attached| is asserted, enabling the subsequent drive-assisted-train action.

\begin{table*}[!t]
\centering
\caption{Disruption-recovery action schemas. Shorthands as in
Table~\ref{tab:actions_movement}. The
$\mathsf{DriveEngineToDamagedDownTrain}$ variant mirrors the up-variant,
with $\mathit{down}$ and $\mathit{heading\_down}$ in place of
$\mathit{up}$ and $\mathit{heading\_up}$, and the engine heading flipped
accordingly. Dispatch actions do not require the destination to be free,
as it is occupied by the disabled train.}
\label{tab:actions_recovery}
\footnotesize
\setlength{\tabcolsep}{5pt}
\begin{tabular}{l l l l l l}
\toprule
\textbf{Action} & \multicolumn{2}{l}{\textbf{Preconditions}}
                & \multicolumn{2}{l}{\textbf{Effects}}
                & \textbf{Duration} \\
\cmidrule(lr){2-3}\cmidrule(lr){4-5}
 & ann. & predicate & ann. & predicate & \\
\midrule
$\mathsf{DriveEngineTo}$
 & Start & $\mathit{engine\_at}(e,n_i)$
 & Start & $\neg\mathit{engine\_at}(e,n_i)$
 & $\tau_{v_a(e)}(n_i,n_j)$ \\
$\mathsf{DamagedUpTrain}$
 & & $\mathit{engine\_free}(e)$
 & & $\mathit{free}(n_i)$ & \\
$(e,t,n_i,n_j,g)$
 & & $\mathit{engine\_gauge}(e,g)$
 & & $\neg\mathit{track\_accessible}(n_i,n_j)$ & \\
 & & $\mathit{track\_gauge}(n_i,n_j,g)$
 & & $\neg\mathit{track\_accessible}(n_j,n_i)$ & \\
 & & $\mathit{train\_gauge}(t,g)$ & & & \\
 & & $\mathit{train\_at}(t,n_j)$ & & & \\
 & & $\mathit{engine\_damaged}(t,n_j)$ & & & \\
 & & $\mathit{up}(n_i,n_j)$ & & & \\
 & & $\mathit{heading\_up}(t)$ & & & \\
 & & $\mathit{heading\_down}(e)$ & & & \\
 & & $\mathit{track\_clear}(n_i,n_j)$ & & & \\
 & & $\mathit{track\_accessible}(n_i,n_j)$ & & & \\
\addlinespace[5pt]
 & Over-all & $\mathit{connected}(n_i,n_j)$
 & End & $\mathit{engine\_at}(e,n_j)$ & \\
 & & $\mathit{train\_at}(t,n_j)$
 & & $\mathit{heading\_up}(e)$ & \\
 & & & & $\neg\mathit{heading\_down}(e)$ & \\
 & & & & $\mathit{track\_accessible}(n_i,n_j)$ & \\
 & & & & $\mathit{track\_accessible}(n_j,n_i)$ & \\
\midrule
$\mathsf{AttachEngine}$
 & Start & $\mathit{engine\_at}(e,n)$
 & Start & $\neg\mathit{engine\_free}(e)$
 & $\mu(e)$ \\
$(e,t,n)$
 & & $\mathit{train\_at}(t,n)$ & & & \\
 & & $\mathit{engine\_free}(e)$
 & End & $\mathit{engine\_attached}(e,t)$ & \\
 & & $\mathit{engine\_damaged}(t,n)$ & & & \\
\addlinespace[5pt]
 & Over-all & $\mathit{engine\_at}(e,n)$ & & & \\
 & & $\mathit{train\_at}(t,n)$ & & & \\
\midrule
$\mathsf{DriveAssistedTrain}$
 & Start & $\mathit{engine\_attached}(e,t)$
 & Start & $\neg\mathit{train\_at}(t,n_i)$
 & $\tau_{v_a(e)}(n_i,n_j)$ \\
$(e,t,n_i,n_j,g)$
 & & $\mathit{train\_at}(t,n_i)$
 & & $\neg\mathit{engine\_at}(e,n_i)$ & \\
 & & $\mathit{engine\_at}(e,n_i)$
 & & $\mathit{free}(n_i)$ & \\
 & & $\mathit{engine\_gauge}(e,g)$
 & & $\neg\mathit{engine\_damaged}(t,n_i)$ & \\
 & & $\mathit{track\_gauge}(n_i,n_j,g)$
 & & $\neg\mathit{track\_accessible}(n_i,n_j)$ & \\
 & & $\mathit{track\_clear}(n_i,n_j)$
 & & $\neg\mathit{track\_accessible}(n_j,n_i)$ & \\
 & & $\neg\mathit{train\_blocked}(t,n_i)$ & & & \\
 & & $\mathit{track\_accessible}(n_i,n_j)$ & & & \\
\addlinespace[5pt]
 & Over-all & $\mathit{connected}(n_i,n_j)$
 & End & $\mathit{train\_at}(t,n_j)$ & \\
 & & $\mathit{free}(n_j)$
 & & $\mathit{engine\_at}(e,n_j)$ & \\
\addlinespace[5pt]
 & End & $\mathit{free}(n_j)$
 & & $\neg\mathit{free}(n_j)$ & \\
 & & & & $\mathit{engine\_damaged}(t,n_j)$ & \\
 & & & & $\mathit{track\_accessible}(n_i,n_j)$ & \\
 & & & & $\mathit{track\_accessible}(n_j,n_i)$ & \\
\midrule
$\mathsf{ResolveTrainBlockage}$
 & Start & $\mathit{train\_blocked}(t,n)$
 & End & $\neg\mathit{train\_blocked}(t,n)$
 & $\rho(t,n)$ \\
$(t,n)$
 & & $\mathit{train\_at}(t,n)$ & & & \\
\addlinespace[5pt]
 & Over-all & $\mathit{train\_at}(t,n)$ & & & \\
\midrule
$\mathsf{ClearBlockedTrack}$
 & Start & $\mathit{track\_blocked}(n_i,n_j)$
 & End & $\mathit{track\_clear}(n_i,n_j)$
 & $\kappa(n_i,n_j)$ \\
$(n_i,n_j)$
 & & & & $\mathit{track\_clear}(n_j,n_i)$ & \\
 & & & & $\neg\mathit{track\_blocked}(n_i,n_j)$ & \\
 & & & & $\neg\mathit{track\_blocked}(n_j,n_i)$ & \\
\bottomrule
\end{tabular}
\end{table*}

Once attached, the \verb|drive-assisted-train| action, shown in Table~\ref{tab:actions_recovery}, enables the damaged train to move, with the auxiliary engine’s speed determining the duration. The original train engine remains non-functional, which is reflected at the destination.

Operational disruptions are resolved using two dedicated recovery actions. The \verb|resolve-train-blockage| action restores mobility of blocked trains, while \verb|clear-blocked-track| models infrastructure restoration following track obstructions.

\subparagraph{Infrastructure Operations}
Infrastructure management primarily involves the \verb|turnout| action (Table \ref{tab:actions_infra}), which switches routing at junction points. The active connection is severed at initiation, while the alternative is established at completion, modeling physical switching delays via \verb|turnout-time|.

\begin{table}[htbp]
\centering
\caption{Infrastructure-operation action schema. $n_1$, $n_2$ denote the
alternative continuation points of the turnout at junction $n$; while the
switching is in progress, neither alternative is connected.}
\label{tab:actions_infra}
\footnotesize
\setlength{\tabcolsep}{5pt}
\begin{tabular}{l l l l l l}
\toprule
\textbf{Action} & \multicolumn{2}{l}{\textbf{Preconditions}}
                & \multicolumn{2}{l}{\textbf{Effects}}
                & \textbf{Duration} \\
\cmidrule(lr){2-3}\cmidrule(lr){4-5}
 & ann. & predicate & ann. & predicate & \\
\midrule
$\mathsf{Turnout}$
 & Start & $\mathit{junction\_point}(n)$
 & Start & $\neg\mathit{connected}(n,n_1)$
 & $\theta$ \\
$(n,n_1,n_2)$
 & & $\mathit{turnout\_alternatives}(n,n_1,n_2)$
 & & $\neg\mathit{connected}(n_1,n)$ & \\
 & & $\mathit{connected}(n,n_1)$
 & End & $\mathit{connected}(n,n_2)$ & \\
 & & $\neg\mathit{connected}(n,n_2)$
 & & $\mathit{connected}(n_2,n)$ & \\
\bottomrule
\end{tabular}
\end{table}
Together, these durative actions provide an integrated temporal representation of railway movement, service operations, infrastructure management, and disruption recovery, enabling the planner to generate executable and disruption-resilient railway schedules.

\paragraph{Temporal resource constraints}
Track segments and points are exclusive temporal resources. 
When a vehicle traverses $(n_i,n_j)$ starting at time $t_s$, the segment is unavailable to all other vehicles during $[t_s,\ t_s+\tau(n_i,n_j)]$, where $\tau$ is the traversal duration of Eq.~\eqref{eq:travelling_time_slowdown}.
The turnout operations hold exclusive access to their junction for the switching interval $turnout-time$. 
These constraints ensure collision-free plans.

\subsubsection{Problem Instance Modeling} 
Each problem instance instantiates $\langle I_R, G_R \rangle$ for a concrete scenario over $R$: the initial state $I_R \subseteq F_R$, together with an assignment to the fluents $X_R$, fixes train positions and gauges, the network topology and segment gauges, junction alternatives, platform and boarding assignments, distances, speeds, and service times, the goal follows specification presented in subsection~\ref{sec:goal_specification}.
Table~\ref{tab:instance_example} presents the instance corresponding to the network of Figure~\ref{fig:railway_network1}.

\begin{table}[!ht]
\centering
\caption{Planning problem for the representative nominal instance of Figure~\ref{fig:railway_network1}, instantiating the objects, initial state ($I_R$), and goal ($G_R$). Symmetric $\mathit{connected}$, $\mathit{track\_accessible}$, and
$\mathit{track\_gauge}$ facts.}
\label{tab:instance_example}
\footnotesize
\setlength{\tabcolsep}{4pt}
\renewcommand{\arraystretch}{1.25}
\begin{tabular}{l l p{6.2cm}}
\toprule
\textbf{Category} & & \textbf{Predicates / values} \\
\midrule
\multirow{3}{*}{Objects}
 & Track points & $A, B, C, D, E, F$ \\
 & Trains & $t_1$ (broad gauge), $t_2$ (meter gauge) \\
 & Stations & $s_1$ \\
\midrule
\multirow{7}{*}{\makecell[l]{Initial\\state ($I_R$)}}
 & Deployment & $\mathit{train\_at}(t_1,A)$, $\mathit{train\_at}(t_2,F)$ \\
 & Topology & $\mathit{connected}(A,B)$, $\mathit{connected}(B,C)$,
   $\mathit{connected}(B,D)$, $\mathit{connected}(C,E)$,
   $\mathit{connected}(D,E)$, $\mathit{connected}(E,F)$ \\
 & Junctions & $\mathit{junction\_point}(B)$,
   $\mathit{turnout\_alternatives}(B,C,D)$,
   $\mathit{junction\_point}(E)$,
   $\mathit{turnout\_alternatives}(E,C,D)$ \\
 & Gauges & $\mathit{train\_gauge}(t_1,\mathit{broad})$,
   $\mathit{train\_gauge}(t_2,\mathit{meter})$,
   $\mathit{track\_gauge}(A,B,\mathit{broad})$, \dots,
   $\mathit{track\_gauge}(E,F,\mathit{meter})$ \\
 & Service & $\mathit{platform\_at}(C,s_1)$,
   $\mathit{platform\_at}(D,s_1)$,
   $\mathit{boarding\_point}(t_1,D)$,
   $\mathit{boarding\_point}(t_2,C)$ \\
 & Static values & $d(A,B){=}60$, $d(B,C){=}d(B,D){=}10$,
   $d(C,E){=}d(D,E){=}15$, $d(E,F){=}40$,
   $v(t_1){=}1.0$, $v(t_2){=}1.1$,
   $\beta(t_1,s_1){=}2$, $\beta(t_2,s_1){=}3$, $\theta{=}1$ \\
\midrule
\multirow{2}{*}{Goal ($G_R$)}
 & Destinations & $\mathit{train\_at}(t_1,F)$, $\mathit{train\_at}(t_2,A)$ \\
 & Service & $\mathit{passengers\_boarded}(t_1,D)$,
   $\mathit{passengers\_boarded}(t_2,C)$ \\
\midrule
Metric & & minimize $\mathit{total\text{-}time}$ (makespan) \\
\bottomrule
\end{tabular}
\end{table}

with metric $\min\,\textit{total-time}$ (makespan minimization). 
Together, the domain model and problem instances encode railway scheduling and routing as a temporal planning task, from which temporal planners automatically generate timestamped action sequences that respect safety and gauge constraints, handle disruptions, and minimize makespan in heterogeneous multi-gauge networks.

\subsection{Temporal Railway Planning Benchmark Generation}
\label{sec:benchmark_generation}
Building on this formulation, we constructed a benchmark comprising a single domain model and 200 problem instances, evenly divided into 100 nominal and 100 disrupted instances. Every instance models heterogeneous railway infrastructure with three track types (meter, broad, and dual-gauge) and two train types (meter and broad-gauge) operating at different speeds, thereby capturing the gauge compatibility and speed heterogeneity of real railway systems. Table~\ref{tab:gen_params} summarizes the fixed parameters used during benchmark generation, including network characteristics, train movement properties, disruption settings, and operational action durations. The benchmark generation procedures are presented in Algorithms~\ref{alg:nominal} and~\ref{alg:disrupt}.

\begin{table}[!htbp]
\centering
\caption{Fixed parameters and sampling ranges. $\mathcal{U}(a,b)$: continuous uniform.}
\label{tab:gen_params}
\small
\begin{tabular}{p{5.6cm} p{3.0cm}}
\toprule
\textbf{Parameter} & \textbf{Value / range} \\
\midrule
Gauge types & \{broad, meter\} \\
Train speed (km/min) & $\mathcal{U}(1.0,\,1.2)$ \\
Auxiliary-engine speed (km/min) & $1.0$ \\
Railway segment length (km) & $\mathcal{U} (15,35)$ \\
Disrupted slowdown factor & $\mathcal{U}(0.15,\,0.30)$ \\
Blocked track clear time (min) & $\mathcal{U} (3,\,60)$ \\
Train blockage recovery time (min) & $\mathcal{U} (2,\,60)$ \\
Turnout (track switching) time (min) & $1$\\
boarding time (min) & $\mathcal{U} (1,5)$ \\
Engine attach time (min) & $\mathcal{U} (1,15)$ \\
\bottomrule
\end{tabular}
\end{table}

\textsc{GenerateNominal} function (Algorithm~\ref{alg:nominal}) takes the target numbers of trains, stations, track points, and junctions together with a random seed as inputs. Then it  constructs a connected railway network, assigns stations, junctions, and trains, and generates the nominal planning task by producing the initial state $I_R$ and goal state $G_R$. The generation process guarantees that every train has at least one gauge-compatible route between its origin and destination, ensuring that every nominal instance is solvable. Nominal-operation instances evaluate planning performance under normal operating conditions. Their complexity is determined solely by the railway infrastructure, with increasing numbers of trains, stations, track points, and junctions. 

\begin{algorithm}[!htbp]
\caption{Nominal railway instance generation.}
\label{alg:nominal}
\begin{algorithmic}[1]
\Function{GenerateNominal}{$n_{tr}, n_{st}, n_{pt}, n_{jc}, \sigma$}
\Statex \textbf{Input:} Number of trains ($n_{tr}$), stations ($n_{st}$), track points ($n_{pt}$), junctions ($n_{jc}$), random seed ($\sigma$)
\Statex \textbf{Output:} Nominal initial state $I_R$ and goal $G_R$
\State Initialize an empty railway network using random seed $\sigma$.
\State Construct a connected dual-gauge corridor consisting of $m$ track points.
\State Attach $n_{st}$ stations to randomly selected corridor points using gauge-compatible tracks.
\State Insert $n_{jc}$ junctions to create alternative parallel routes.
\State Assign $n_{tr}$ trains to randomly selected gauge-compatible origin and destination stations.
\State Initialize track attributes and construct the connectivity and gauge relations.
\State Construct the nominal initial state $I_R$ and goal state $G_R$.
\State \Return $\langle I_R,G_R\rangle$
\EndFunction
\end{algorithmic}
\end{algorithm}

On the other hand, \textsc{GenerateDisrupted} (Algorithm~\ref{alg:disrupt}) starts from a nominal instance and introduces the specified numbers of blocked trains, blocked track segments, slowdown-affected segments, and engine failures to generate a disrupted initial state $I_R'$, while preserving the original goal state $G_R$. During disruption generation, track blockages are placed without disconnecting the railway network, ensuring that every disrupted instance remains solvable. Disrupted-operation instances evaluate planning performance under both increasing infrastructure size and disruption severity. 

\begin{algorithm}[!htbp]
\caption{Disrupted railway instance generation.}
\label{alg:disrupt}
\begin{algorithmic}[1]
\Function{GenerateDisrupted}{$I_R,n_{bt},n_{btr},n_{sd},n_{ef}$}
\Statex \textbf{Input:} Nominal initial state $I_R$, blocked trains ($n_{bt}$), blocked tracks ($n_{btr}$), slowdowns ($n_{sd}$), engine failures ($n_{ef}$)
\Statex \textbf{Output:} Disrupted initial state $I_R'$
\State Select $n_{bt}$ trains and assign blockage durations.
\State Introduce $n_{btr}$ blocked track segments while preserving route connectivity for every train.
\State Apply speed reductions to $n_{sd}$ randomly selected unblocked track segments.
\State Introduce $n_{ef}$ engine failures and assign a reachable rescue engine to each affected train.
\State Update the railway state with the generated disruptions.
\State \Return $I_R'$
\EndFunction
\end{algorithmic}
\end{algorithm}

The benchmark instances are organized into four primary scale categories: Small (S), Medium (M), Large (L), and Very Large (VL). Each category represents a progressive increase in infrastructure size and planning complexity. Within each category, multiple sub-levels (e.g., S1--S2, M1--M2, L1--L3, and VL1--VL3) provide finer-grained variations in problem difficulty. Tables~\ref{tab:problem_set1} and~\ref{tab:problem_set2} summarize the characteristics of the nominal and disrupted benchmark instances, respectively.

\begin{table}[htbp!]
\caption{Nominal-operation benchmark instances grouped by increasing problem size. 
Instance complexity progressively increases with respect to the number of trains, 
stations, track points, and junctions.}
\centering
\renewcommand{\arraystretch}{1.2}
\begin{tabular}{clcccc}
\toprule
\textbf{Scale} & \textbf{Instances} & \textbf{Trains} & \textbf{Stations} & \textbf{Track Points} & \textbf{Junctions} \\
\midrule
S1 & p1--p10   & 3--12    & 5--17     & 50--127   & 2--7   \\
S2 & p11--p20  & 14--23   & 18--31    & 136--213  & 8--13  \\
\midrule
M1 & p21--p30  & 25--35   & 32--45    & 222--300  & 14--20 \\
M2 & p31--p40  & 35--45   & 45--57    & 300--392  & 20--29 \\
\midrule
L1 & p41--p50  & 46--56   & 59--71    & 402--494  & 30--39 \\
L2 & p51--p60  & 58--68   & 73--85    & 505--597  & 40--49 \\
L3 & p61--p70  & 69--80   & 87--100   & 607--700  & 50--60 \\
\midrule
VL1 & p71--p80  & 80--92   & 100--115  & 700--793  & 60--72 \\
VL2 & p81--p90  & 93--106  & 117--132  & 803--896  & 73--86 \\
VL3 & p91--p100 & 107--120 & 134--150  & 906--1000 & 87--100 \\
\bottomrule
\end{tabular}
\label{tab:problem_set1}
\end{table}

\begin{sidewaystable}[htbp]
\caption{Disrupted-operation benchmark instances grouped by increasing 
problem and disruption complexity. In addition to infrastructure size, 
instances vary in disruption severity including blocked trains, blocked 
tracks, engine failures, slowdown segments, and available auxiliary engines.}
\centering
\renewcommand{\arraystretch}{1.2}
\setlength{\tabcolsep}{4pt}

\begin{tabular}{clccccccccc}
\toprule
\multirow{3}{*}{\textbf{Scale}} & \multirow{3}{*}{\textbf{Instances}} &
\multicolumn{4}{c}{\textbf{Problem Characteristics}} &
\multicolumn{5}{c}{\textbf{Disruption Characteristics}} \\
\cmidrule(lr){3-6}\cmidrule(lr){7-11}

& & \textbf{Trains} & \textbf{Stations} & \textbf{Track Points} & \textbf{Junctions}
& \textbf{Blocked} & \textbf{Blocked} & \textbf{Engine} & \textbf{Slow-} & \textbf{Aux.} \\
& & & & & 
& \textbf{Train} & \textbf{Track} & \textbf{Failure} & \textbf{down} & \textbf{Eng.} \\

\midrule

S1 & p101--p110 & 3--12   & 5--17   & 51--128   & 2--7   & 1--1 & 1--3  & 1--1 & 1--3  & 1--1 \\
S2 & p111--p120 & 14--23  & 18--31  & 137--214  & 8--13  & 1--1 & 3--4  & 1--1 & 3--5  & 1--1 \\

\midrule

M1 & p121--p130 & 25--35  & 32--45  & 223--301  & 14--20 & 1--2 & 5--6  & 1--1 & 6--8  & 1--1 \\
M2 & p131--p140 & 35--45  & 45--57  & 302--395  & 20--29 & 2--3 & 10--13& 2--3 & 13--17& 2--3 \\

\midrule

L1 & p141--p150 & 46--56  & 59--71  & 405--497  & 30--39 & 3--4 & 13--16& 3--3 & 17--21& 3--3 \\
L2 & p151--p160 & 58--68  & 73--85  & 508--601  & 40--49 & 4--5 & 16--19& 3--4 & 22--26& 3--4 \\
L3 & p161--p170 & 69--80  & 87--100 & 611--705  & 50--60 & 5--6 & 20--23& 4--5 & 26--30& 4--5 \\

\midrule

VL1 & p171--p180 & 80--92   & 100--115 & 706--800  & 60--72 & 8--9  & 30--34 & 6--7 & 46--51 & 6--7 \\
VL2 & p181--p190 & 93--106  & 117--132 & 810--904  & 73--86 & 9--10 & 34--38 & 7--8 & 52--57 & 7--8 \\
VL3 & p191--p200 & 107--120 & 134--150 & 914--1008 & 87--100&10--10 & 38--42 & 8--8 & 57--63 & 8--8 \\

\bottomrule
\end{tabular}
\label{tab:problem_set2}
\end{sidewaystable}

The benchmark is designed with controlled and monotonic scaling. As the scale increases from S to VL, the numbers of trains, stations, track points, and junctions increase consistently. For disrupted instances, disruption severity grows in parallel with infrastructure size through increasing numbers of blocked trains, blocked tracks, slowdown-affected segments, engine failures, and auxiliary engines. This design enables systematic evaluation of planner scalability under progressively larger railway networks and increasingly challenging disruption scenarios.

\section{Experimental Results}
\label{sec:ex_results}

This section presents a comprehensive experimental evaluation of the proposed temporal planning framework. We evaluate the railway domain and benchmark problem set using the temporal planners POPF3~\cite{coles2011popf2} and OPTIC~\cite{benton2012temporal}. Both planners are satisficing planners that terminate after finding the first valid solution rather than attempting to compute an optimal plan. Consequently, all plan quality values reported in this section correspond to the quality of the generated (achieved) plans and should not be interpreted as optimal. To ensure correctness, we also validated all generated temporal plans using the VAL plan validator~\cite{VAL_1374201}, which verifies their semantic consistency with the PDDL domain specification.

The two planners (POPF and OPTIC) used in this experiment employ different search mechanisms and heuristic techniques for solving temporal planning problems. 
POPF uses grounded forward search with linear programming, and a Temporal Relaxed Planning Graph (TRPG) heuristic to guide exploration. 
OPTIC extends POPF with mixed-integer programming techniques for optimization, allowing preference-aware temporal planning, and employs an enhanced TRPG heuristic with improved pruning during search.
Both planners were executed with their default parameter settings and a maximum time limit of 30 minutes per problem instance. 
Since both planners operate deterministically, each problem instance was solved once. 
All experiments were conducted on a workstation equipped with a 5th-generation Intel Core i5 processor, and 8\,GB RAM, running Ubuntu 24.04.3 LTS.

To evaluate the generated temporal plans, we consider several standard planning metrics: \emph{makespan} (Definition~\ref{def:makespan}), \emph{plan length}, and \emph{plan generation time}. Makespan measures the total execution time of a plan, plan length denotes the total number of actions in the generated operational plan, and plan generation time is the wall-clock time required by a planner to produce a valid solution. Although these metrics assess the efficiency of the planning process, they do not directly capture the operational impact of disruptions on railway performance. We therefore complement them with domain-specific delay metrics that quantify the deviation of the generated schedules from ideal uninterrupted train operations.

\paragraph{Delay Metrics}
To assess the effectiveness of the proposed temporal planning model under realistic operational conditions, we analyze the delays incurred by trains during plan execution. Let $\Pi_T$ denote a temporal plan, $\mathcal{T}={1,\ldots,n}$ the set of trains, and $\mathcal{J}_i={1,\ldots,m_i}$ the ordered sequence of track points traversed by train $i$ in $\Pi_T$. We first define the overall schedule delay and then decompose it into domain-specific components corresponding to different disruption types. Total Delay (Definition~\ref{def:total_delay}) quantifies overall schedule degradation under operational conditions.
\begin{definition}[Total Delay]\label{def:total_delay}
The \textbf{total delay} of a temporal plan $\Pi_T$ is the cumulative deviation between the actual and ideal arrival times of all trains at all track points:
\[
\mathrm{TotalDelay}(\Pi_T)=
\sum_{i=1}^{n}\sum_{j=1}^{m_i}
\left(T_{ij}^{\mathrm{actual}}-T_{ij}^{\mathrm{ideal}}\right),
\]
where $T_{ij}^{\mathrm{actual}}$ and $T_{ij}^{\mathrm{ideal}}$ denote the actual and ideal arrival times of train $i$ at track point $j$, respectively. The ideal arrival time assumes uninterrupted travel at nominal speed:
\[
T_{ij}^{\mathrm{ideal}}
=
T_{i0}^{\mathrm{dep}}
+
\sum_{k=1}^{j}\tau_{ik},
\]
where $T_{i0}^{\mathrm{dep}}$ is the departure time and $\tau_{ik}$ is the minimum traversal time of segment $k$.
\end{definition}

Although total delay measures overall schedule degradation, it does not identify the underlying causes. 
We therefore decompose it into three components, each defined with respect to actions of our railway domain: slowdown delay (Definition~\ref{def:slowdown-delay}), blockage delay (Definition~\ref{def:blockage-delay}), and engine failure delay (Definition~\ref{def:engine-failure-delay}).

\begin{definition}\label{def:slowdown-delay}
The \textbf{slowdown delay} is the extra time incurred when trains traverse segments with reduced speed. For each \texttt{drive-train} or \texttt{drive-assisted-train} action on segment $(p,q)$, let $\tau^{\mathrm{normal}} = d_{pq}/s_i$ be the nominal duration and $\tau^{\mathrm{slow}} = d_{pq}/((1-\alpha_{pq}) s_i)$ the actual duration under slowdown factor $\alpha_{pq} \in [0,1)$. Then:
\[
\mathrm{SlowdownDelay}(\Pi_T) = 
\sum_{\substack{a \in \Pi_T \\ a \in \{\texttt{drive-train},\, \texttt{drive-assisted-train}\}}} 
\bigl(\mathrm{dur}(a) - \tau^{\mathrm{normal}}\bigr),
\]
where the difference is zero if $\alpha_{pq}=0$.
\end{definition}

\begin{definition}\label{def:blockage-delay}
The \textbf{blockage delay} is the total waiting time due to train or track blockages, computed as the cumulative duration of all \texttt{resolve-train-blockage} and \texttt{clear-blocked-track} actions:
\[
\mathrm{BlockageDelay}(\Pi_T) = 
\sum_{\substack{a \in \Pi_T \\ a = \texttt{resolve-train-blockage}}} 
\mathrm{dur}(a)
+
\sum_{\substack{a \in \Pi_T \\ a = \texttt{clear-blocked-track}}} 
\mathrm{dur}(a),
\]
where each action duration is defined in the domain.
\end{definition}

\begin{definition}\label{def:engine-failure-delay}
The \textbf{engine failure delay} is the total immobilization time of trains experiencing engine failure. For each train $i \in \mathcal{F}$, recovery involves dispatching an auxiliary engine, attaching it, and resuming movement. The delay is the sum of the dispatch and attach durations:
\[
\mathrm{EngineFailureDelay}(\Pi_T) = 
\sum_{i \in \mathcal{F}} 
\bigl( \mathrm{dur}(a_i^{\mathrm{dispatch}}) + \mathrm{dur}(a_i^{\mathrm{attach}}) \bigr),
\]
where durations are defined by the corresponding domain actions (\texttt{drive-engine}, \texttt{drive-engine-to-damaged-train} and \texttt{attach-engine}).
\end{definition}

By decomposing total delay into slowdown, blockage, and engine failure components with respect to our domain, this study provides a detailed assessment of how different types of disruptions affect the generated plans.

\subsection{Overall Experimental Results}
Tables~\ref{tab:normal_results} summarize the performance of the OPTIC and POPF temporal planners on the proposed railway domain and benchmark problem set.  Both planners solved all 100 nominal instances and 99 of the 100 disrupted ones. 
And produced identical makespan, plan length, and delay values on every commonly solved instance, a single unified set of quality results is therefore reported. 
We attribute this convergence to OPTIC extending POPF with the same underlying TRPG heuristic and search strategy, and to the tightly constrained solution space imposed by gauge compatibility and single-track mutual exclusion, which leaves both planners little room to diverge in the plans they construct.

\begin{table}[!htbp]
\centering
\caption{Experimental results of OPTIC and POPF on nominal and
disrupted-operation instances. Makespan and delay are reported in minutes and
plan length in actions, as mean $\pm$ standard deviation over ten instances per
group. Both planners produced plans of identical quality on every commonly
solved instance; a single unified set of quality metrics is therefore reported.}
\label{tab:normal_results}
\begin{tabular}{l rrrrrr}
\toprule
\multirow{2}{*}{\textbf{Scale}}
  & \multicolumn{3}{c}{\textbf{Nominal-operation instances}}
  & \multicolumn{3}{c}{\textbf{Disrupted-operation instances}} \\
\cmidrule(lr){2-4} \cmidrule(lr){5-7}
& Makespan & Plan Length & Delay
& Makespan & Plan Length & Delay \\
\midrule
S1   & $30.5{\pm}3.1$ & $15.0{\pm}6.1$  & $0.0$ & $42.6{\pm}19.4$ & $18.0{\pm}6.1$  & $28.0{\pm}42.2$   \\
S2   & $33.9{\pm}1.4$ & $37.0{\pm}6.1$  & $0.0$ & $46.3{\pm}18.8$ & $40.0{\pm}6.1$  & $44.1{\pm}92.8$   \\
\midrule
M1   & $33.8{\pm}0.8$ & $59.2{\pm}6.4$  & $0.0$ & $46.7{\pm}25.4$ & $62.7{\pm}6.9$  & $60.8{\pm}140.9$  \\
M2   & $34.5{\pm}1.3$ & $79.6{\pm}6.9$  & $0.0$ & $49.1{\pm}14.5$ & $86.7{\pm}7.4$  & $88.5{\pm}176.5$  \\
\midrule
L1   & $34.8{\pm}1.5$ & $102.4{\pm}6.9$ & $0.0$ & $50.0{\pm}25.1$ & $112.2{\pm}7.2$ & $120.3{\pm}229.6$ \\
L2   & $34.5{\pm}0.6$ & $125.8{\pm}7.0$ & $0.0$ & $52.0{\pm}20.2$ & $138.2{\pm}7.9$ & $140.4{\pm}254.5$ \\
L3   & $34.6{\pm}0.9$ & $148.6{\pm}7.2$ & $0.0$ & $53.0{\pm}24.0$ & $163.3{\pm}8.5$ & $170.3{\pm}317.3$ \\
\midrule
VL1  & $35.0{\pm}0.7$ & $171.6{\pm}8.3$ & $0.0$ & $54.2{\pm}25.7$ & $193.2{\pm}9.4$ & $217.1{\pm}358.4$ \\
VL2  & $35.2{\pm}1.0$ & $199.0{\pm}8.5$ & $0.0$ & $55.0{\pm}29.6$ & $224.1{\pm}9.6$ & $255.9{\pm}425.3$ \\
VL3$^{\ast}$ & $35.4{\pm}0.7$ & $226.6{\pm}8.6$ & $0.0$ & $74.6{\pm}16.2$ & $251.1{\pm}7.7$ & $655.8{\pm}490.4$ \\
\midrule
\textbf{Mean}
     & $\mathbf{34.2{\pm}1.9}$ & $\mathbf{116.5{\pm}67.4}$ & $\mathbf{0.0}$
     & $\mathbf{52.1{\pm}22.8}$ & $\mathbf{127.7{\pm}75.0}$ & $\mathbf{173.3{\pm}319.8}$ \\
\bottomrule
\multicolumn{7}{l}{\footnotesize $^{\ast}$Disrupted VL3 statistics are computed
over the nine solved instances.}
\end{tabular}
\end{table}

For nominal-operation instances, makespan remains stable across all scales with low standard deviations ($34.2 \pm 1.9$ minutes overall), plan length grows proportionally with network size, and total delay is exactly zero, confirming conflict-free scheduling under undisrupted conditions. 
For disrupted-operation instances, makespan, plan length, and delay all increase with problem scale.
Plan-length growth reflects the added recovery operations, while total delay exhibits the most pronounced increase, driven by the escalating number and diversity of disruptions. 
The high standard deviations, particularly in delay, reflect inter-instance variability in disruption severity, and rerouting availability, which we examine in Section~\ref{sec:delay_analysis}.

The single failure, common to both planners, was the largest configuration in the benchmark (instance p200: 120 trains, 1{,}000 track points, and 108 concurrent disruptions), on which both planners exhausted the available physical memory of 8GB and were terminated by the operating system before reaching the 30-minute time limit, no solution had been emitted at that point.
Consequently, the disrupted VL3 statistics in Table~\ref{tab:normal_results} are computed over its nine solved instances. 

All 199 generated plans were verified with the VAL validator~\cite{VAL_1374201}, which confirmed that every plan satisfies the temporal constraints, precondition requirements, and goal conditions specified in the domain.
No mutex violations, precondition failures, or goal-achievement errors were detected. 
This establishes the semantic correctness of the plans \emph{with respect to the PDDL domain encoding}.
Therefore, the VAL validation certifies the internal consistency of the generated recovery schedules for the encoded railway model, while deployment in practice would require an additional conformance layer between the planned schedule and the operational traffic-management system.

\subsection{Delay Analysis}
\label{sec:delay_analysis}
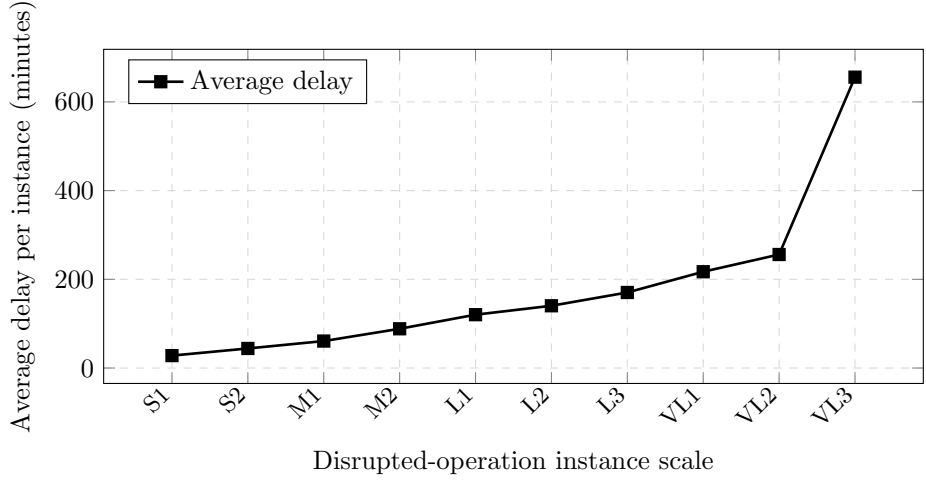
\begin{figure}[!htbp]
\centering
\begin{tikzpicture}
\begin{axis}[
    width=0.95\columnwidth,
    height=6cm,
    xlabel={Disrupted-operation instance scale},
    ylabel={Average delay per instance (minutes)},
    xticklabels={S1, S2, M1, M2, L1, L2, L3, VL1, VL2, VL3},
    xtick={1,2,3,4,5,6,7,8,9,10},
    x tick label style={rotate=45, anchor=east, font=\small},
    legend pos=north west,
    grid=major,
    grid style={dashed, gray!30},
    every axis plot/.append style={thick},
    mark size=2pt,
]
\addplot[black, mark=square*, line width=1pt] coordinates {
    (1, 28.00) (2, 44.11) (3, 60.78) (4, 88.48) (5, 120.27)
    (6, 140.36) (7, 170.32) (8, 217.10) (9, 255.93) (10, 655.79)
};
\addlegendentry{Average delay}
\end{axis}
\end{tikzpicture}
\caption{Average delay per instance for disrupted-operation instances (VL3 averaged over its nine solved instances). Both OPTIC and POPF generate identical coordination protocols with the same delay values. Delay grows near-linearly up to VL2 and surges at VL3, driven primarily by blockage-resolution delay (cf.\ Figure~\ref{fig:delay_breakdown}).}
\label{fig:total_delay}
\end{figure}

No delay was observed on any nominal-operation instance, confirming that the generated schedules are conflict-free in the absence of disruptions.
Figure~\ref{fig:total_delay} shows the average total delay across the disrupted-instance groups. 
Delay grows steadily and near-linearly with problem scale from S1 (28.0 minutes) to VL2 (255.9 minutes), and then rises sharply at VL3 (655.8 minutes over the nine solved instances). 
OPTIC and POPF produce identical delay values at every scale point, indicating that the domain is constrained tightly enough for both planners to converge on the same solution rather than finding different valid alternatives.

\begin{figure}[!htbp]
\centering
\begin{tikzpicture}
\begin{axis}[
    width=0.95\columnwidth,
    height=6cm,
    ybar stacked,
    xlabel={Disrupted-operation instance scale},
    ylabel={Average delay per instance (minutes)},
    xticklabels={S1, S2, M1, M2, L1, L2, L3, VL1, VL2, VL3},
    xtick={1,...,10},
    x tick label style={rotate=45, anchor=east, font=\small},
    legend pos=north west,
    grid=major,
    grid style={dashed, gray!30},
    bar width=7pt,
]
\addplot[fill=black!20, draw=black, line width=0.5pt] coordinates {
    (1, 2.8) (2, 2.9) (3, 4.3) (4, 8.9) (5, 11.5)
    (6, 14.9) (7, 17.0) (8, 24.2) (9, 28.7) (10, 240.1)
};
\addlegendentry{Blockage delay}
\addplot[fill=black!50, draw=black, line width=0.5pt] coordinates {
    (1, 12.0) (2, 12.0) (3, 12.0) (4, 25.2) (5, 36.0)
    (6, 45.6) (7, 54.0) (8, 80.4) (9, 93.6) (10, 122.8)
};
\addlegendentry{Engine failure delay}
\addplot[fill=black!75, draw=black, line width=0.5pt] coordinates {
    (1, 13.2) (2, 29.2) (3, 44.5) (4, 54.4) (5, 72.8)
    (6, 79.9) (7, 99.3) (8, 112.5) (9, 133.6) (10, 292.9)
};
\addlegendentry{Slowdown delay}
\end{axis}
\end{tikzpicture}
\caption{Breakdown of average delay per instance by disruption type across problem scale. 
Slowdown delay is the main contributor overall (52.6\% of total delay), followed by engine-failure delay (28.1\%) and blockage delay (19.3\%).}
\label{fig:delay_breakdown}
\end{figure}
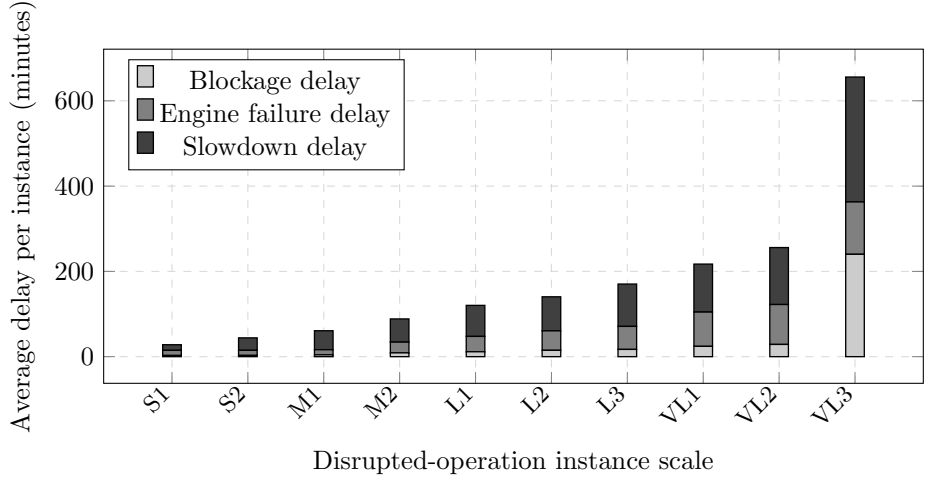

Figure~\ref{fig:delay_breakdown} decomposes total delay by disruption type.
Across all solved instances, slowdowns are the largest source of delay, accounting for 52.6\% of total delay (9{,}029.7 minutes), followed by engine-failure recovery (28.1\%, 4{,}813.0 minutes) and blockage resolution (19.3\%, 3{,}313.0 minutes). 
Up to VL2, blockage delay remains the smallest component throughout (2.8-28.7 minutes per instance), indicating that the planners resolve track conflicts at modest temporal cost, typically by routing around affected segments, without violating temporal constraints. 
The VL3 surge, in contrast, is driven primarily by blockage-resolution delay, which increases $8.4\times$ over VL2 (from 28.7 to 240.1 minutes per instance, 36.6\% of VL3 delay) as the density of simultaneous blockages exhausts the available rerouting options and forces trains to wait for tracks to be cleared.
Slowdown delay also more than doubles ($2.2\times$, to 292.9 minutes per instance, 44.7\% of VL3 delay). 
This distribution confirms that coordination overhead arises from operational constraints encoded in the domain rather than from modeling artifacts, and identifies rerouting capacity as the binding resource at the highest disruption densities.

Table~\ref{tab:delay_distribution} summarizes the delay distribution within each scale group: the first and third quartiles (Q1, Q3) bound the middle 50\% of instances, and Max is the single worst instance in the group. 
The distribution is strongly right-skewed (skewness $2.90$ over all solved instances, overall median 62.0 minutes against a mean of 173.3).
The inter-quartile ranges are narrow (e.g., L1: 46.5-49.8 min), yet one severely disrupted instance per group reaches 5 to 20 times the group median.
In aggregate, the ten most delayed instances account for 58.9\% of all delay in the benchmark, and the top five alone for 39.5\%. 
Combined with the near scale-independence of slowdown delay (Table~\ref{tab:instance_correlations}), this shows that severe degradation depends on which segments are disrupted rather than on network size, while a few unfavourable ones where slowed or blocked segments lie on critical traffic paths dominate system-wide delay. 
This suggests that infrastructure hardening or recovery resources targeted at critical segments would yield disproportionate benefit.

\begin{table}[!htbp]
\centering
\caption{Distribution of total delay (minutes per instance) within each disrupted-operation scale group.}
\label{tab:delay_distribution}
\begin{tabular}{l rrrrr}
\toprule
\textbf{Scale} & \textbf{Mean} & \textbf{Median} & \textbf{Q1} &
\textbf{Q3} & \textbf{Max} \\
\midrule
S1  & 28.0  & 14.5  & 14.0  & 15.8  & 148.0  \\
S2  & 44.1  & 15.0  & 14.2  & 15.0  & 308.1  \\
M1  & 60.8  & 16.5  & 15.2  & 17.8  & 461.8  \\
M2  & 88.5  & 32.5  & 32.0  & 33.8  & 590.8  \\
L1  & 120.3 & 48.0  & 46.5  & 49.8  & 773.7  \\
L2  & 140.4 & 62.5  & 60.5  & 64.8  & 864.6  \\
L3  & 170.3 & 72.5  & 64.0  & 77.8  & 1073.2 \\
VL1 & 217.1 & 106.5 & 99.8  & 108.5 & 1237.0 \\
VL2 & 255.9 & 123.5 & 120.2 & 128.0 & 1466.3 \\
VL3 & 655.8 & 481.0 & 356.0 & 575.0 & 1519.2 \\
\midrule
\textbf{All} & \textbf{173.3} & \textbf{62.0} & \textbf{17.5} &
\textbf{115.0} & \textbf{1519.2} \\
\bottomrule
\end{tabular}
\end{table}

\subsection{Disruption Resilience}
\label{sec:resilience}
Relating disrupted to nominal results quantifies how much of the incorporated disruption the generated recovery schedules absorb (Table~\ref{tab:resilience}).
Three observations follow. 
First, the \emph{makespan inflation factor}, the ratio of disrupted to nominal group makespan, stays between $1.37$ and $1.56$ from S1 through VL2, reaching $2.11$ only at VL3.
Second, the \emph{recovery overhead}, additional actions in the disrupted plan relative to the nominal plan, remains within 6-13\% of nominal plan length above S1, indicating that recovery relies on a modest number of targeted actions rather than rescheduling. 
Third, the \emph{absorption ratio}, total train delay in train-minutes per minute of makespan extension, grows monotonically from $2.3$ at S1 to $16.7$ at VL3. 
Since total delay sums over all trains while makespan reflects only the longest completion, this growth indicates increasingly \emph{parallel} absorption of disruption: trains incur delay simultaneously across separate parts of the network rather than serializing behind a single bottleneck. 
The recovery plans thus exploit the spatial parallelism of larger networks, the behavior required of a practical disruption-management tool.

\begin{table}[!htbp]
\centering
\caption{Resilience metrics per scale group, relating disrupted-operation results to nominal-operation baselines (group means). $\Delta$MS is the increase in mean makespan over nominal, inflation is the disrupted/nominal makespan ratio, overhead is the additional plan length relative to nominal, and absorption is mean total delay (train-minutes) per minute of makespan extension.}
\label{tab:resilience}
\begin{tabular}{l rrrr}
\toprule
\textbf{Scale} & \textbf{Inflation} & \textbf{$\Delta$MS (min)} &
\textbf{Overhead (\%)} & \textbf{Absorption} \\
\midrule
S1  & 1.39 & 12.0 & 20.0 & 2.3  \\
S2  & 1.37 & 12.4 & 8.1  & 3.6  \\
M1  & 1.38 & 12.9 & 5.9  & 4.7  \\
M2  & 1.42 & 14.5 & 8.9  & 6.1  \\
L1  & 1.44 & 15.2 & 9.6  & 7.9  \\
L2  & 1.51 & 17.5 & 9.9  & 8.0  \\
L3  & 1.53 & 18.4 & 9.9  & 9.2  \\
VL1 & 1.55 & 19.3 & 12.6 & 11.3 \\
VL2 & 1.56 & 19.8 & 12.6 & 12.9 \\
VL3 & 2.11 & 39.2 & 10.8 & 16.7 \\
\bottomrule
\end{tabular}
\end{table}

\subsection{Computational Performance}
\label{sec:computation_performance}

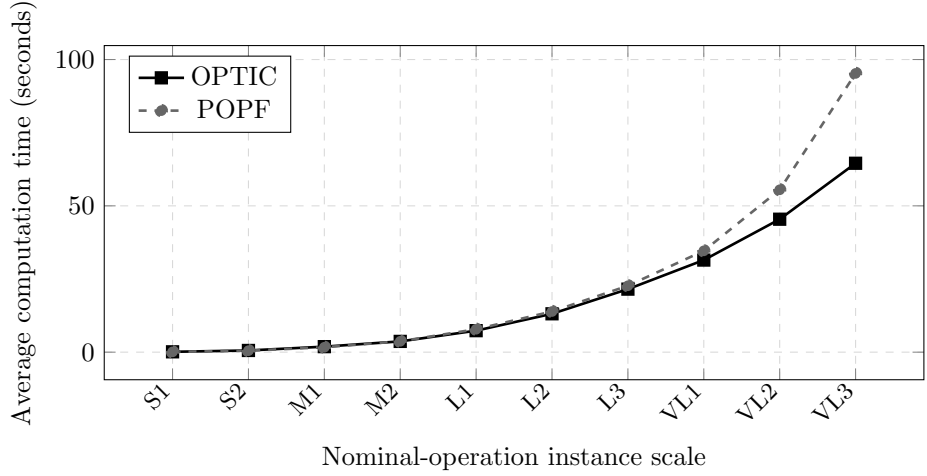
\begin{figure}[!htbp]
\centering
\begin{tikzpicture}
\begin{axis}[
    width=0.95\columnwidth,
    height=6cm,
    xlabel={Nominal-operation instance scale},
    ylabel={Average computation time (seconds)},
    xticklabels={S1, S2, M1, M2, L1, L2, L3, VL1, VL2, VL3},
    xtick={1,2,3,4,5,6,7,8,9,10},
    x tick label style={rotate=45, anchor=east, font=\small},
    legend pos=north west,
    grid=major,
    grid style={dashed, gray!30},
    every axis plot/.append style={thick},
    mark size=2pt,
]
\addplot[black, mark=square*, line width=1pt] coordinates {
    (1, 0.118) (2, 0.579) (3, 1.882) (4, 3.678) (5, 7.347)
    (6, 13.111) (7, 21.537) (8, 31.472) (9, 45.432) (10, 64.555)
};
\addlegendentry{OPTIC}
\addplot[black!60, mark=*, line width=1pt, dashed] coordinates {
    (1, 0.091) (2, 0.503) (3, 1.669) (4, 3.710) (5, 7.829)
    (6, 13.892) (7, 22.689) (8, 34.628) (9, 55.489) (10, 95.257)
};
\addlegendentry{POPF}
\end{axis}
\end{tikzpicture}
\caption{Average computation time per instance for nominal-operation instances.
Both planners scale smoothly and monotonically, POPF is faster up to scale M1 and OPTIC from M2 onwards, with the gap widening to $1.5\times$ at VL3 (64.6\,s against 95.3\,s).}
\label{fig:computation_time_normal}
\end{figure}

\begin{figure}[!htbp]
\centering
\begin{tikzpicture}
\begin{axis}[
    width=0.95\columnwidth,
    height=6cm,
    xlabel={Disrupted-operation instance scale},
    ylabel={Average computation time (seconds)},
    xticklabels={S1, S2, M1, M2, L1, L2, L3, VL1, VL2, VL3},
    xtick={1,2,3,4,5,6,7,8,9,10},
    x tick label style={rotate=45, anchor=east, font=\small},
    legend pos=north west,
    grid=major,
    grid style={dashed, gray!30},
    every axis plot/.append style={thick},
    mark size=2pt,
    ymode=log,
]
\addplot[black, mark=square*, line width=1pt] coordinates {
    (1, 21.828) (2, 3.238) (3, 11.358) (4, 35.243) (5, 72.106)
    (6, 118.841) (7, 123.992) (8, 259.295) (9, 417.329) (10, 625.902)
};
\addlegendentry{OPTIC}
\addplot[black!60, mark=*, line width=1pt, dashed] coordinates {
    (1, 0.282) (2, 2.084) (3, 7.371) (4, 22.640) (5, 48.150)
    (6, 99.568) (7, 129.405) (8, 270.150) (9, 450.420) (10, 639.053)
};
\addlegendentry{POPF}
\end{axis}
\end{tikzpicture}
\caption{Average computation time per instance for disrupted-operation instances (VL3 averaged over its nine solved instances). 
POPF is faster on small instances (0.28 s at S1 against 21.83 s for OPTIC) but its runtime grows more steeply with scale.}
\label{fig:computation_time_disrupted}
\end{figure}
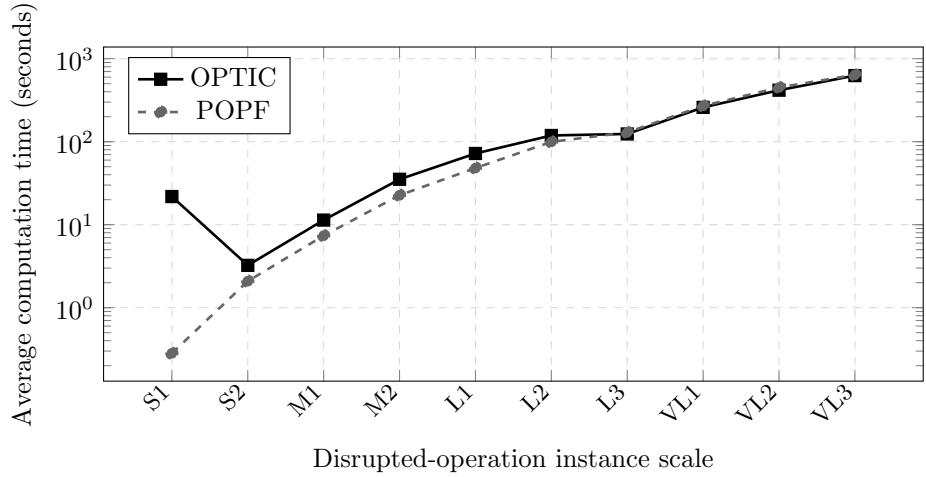

Figures~\ref{fig:computation_time_normal} and \ref{fig:computation_time_disrupted} show the growth of average computation time with problem size. 
Under nominal conditions (Figure~\ref{fig:computation_time_normal}), both planners scale smoothly and monotonically, from 0.09-0.12 seconds at S1 to 64.6 seconds (OPTIC) and 95.3 seconds (POPF) at VL3. 
For disrupted instances (Figure~\ref{fig:computation_time_disrupted}), computation times are substantially higher due to the additional coordination and recovery reasoning, and POPF's scaling exponent rises to $3.01$, quantifying the extra computational burden that disruption handling imposes beyond network growth alone.

Even without disruptions, the two planners differ measurably in efficiency.
OPTIC is faster overall on the nominal instance set (mean 19.0 s against 23.6 s, Wilcoxon signed-rank test $n{=}100$, $z{=}-3.94$, $p<0.001$). 
The same size dependence observed under disruption, at a smaller scale threshold: POPF is faster on 20 of the 21 smallest instances (plan length $\leq 50$), whereas OPTIC is faster on 25 of 45 medium and 28 of 34 large instances.
Their IPC agile scores are correspondingly close (97.47 for OPTIC against 96.54 for POPF), reflecting that neither planner dominates across the whole nominal benchmark.

Since both planners deliver identical plan quality, their comparison reduces to computational efficiency and robustness, summarized in Table~\ref{tab:planner_comparison}. 
We assess efficiency with four measures.
(i)~The \emph{Wilcoxon signed-rank test}, a non-parametric paired test applied to the per-instance runtime differences over the commonly solved instances.
It makes no normality assumption, which is appropriate given the skewed runtime distribution. 
(ii)~The \emph{IPC agile score}, the time-quality measure of the International Planning Competition: each instance contributes $1/(1+\log_{10}(t_i/t_i^{\ast}))$ to a planner's score, where $t_i$ is that planner's runtime and $t_i^{\ast}$ the fastest runtime recorded on instance $i$, unsolved instances contribute $0$, so the maximum score equals the number of instances (here 100) and the fastest planner on an instance receives the full point. 
(iii)~The \emph{geometric mean} of runtimes, which, unlike the arithmetic mean, is not dominated by the largest instances when runtimes span four orders of magnitude.
(iv)~The \emph{runtime scaling exponent} $b$, obtained by least-squares regression of $\log t$ on $\log(\text{plan length})$, i.e., $t \approx c \cdot \mathrm{PL}^{\,b}$; throughout, plan length serves as the instance-scale proxy, since it grows deterministically with network size in the benchmark.

\begin{table}[!htbp]
\centering
\caption{Computational comparison of OPTIC and POPF on the nominal and disrupted benchmark sets. 
Plan quality is identical for both planners on every commonly solved instance in both sets.}
\label{tab:planner_comparison}
\small
\setlength{\tabcolsep}{4pt}
\begin{tabular}{l
    >{\centering\arraybackslash}p{1.6cm}
    >{\centering\arraybackslash}p{1.6cm}
    >{\centering\arraybackslash}p{1.6cm}
    >{\centering\arraybackslash}p{1.6cm}}
\toprule
 & \multicolumn{2}{c}{\textbf{Nominal}} & \multicolumn{2}{c}{\textbf{Disrupted}} \\
\cmidrule(lr){2-3}\cmidrule(lr){4-5}
\textbf{Metric} & \textbf{OPTIC} & \textbf{POPF} & \textbf{OPTIC} & \textbf{POPF} \\
\midrule
Coverage                    & 100/100 & 100/100 & 99/100 & 99/100 \\
IPC agile score             & \textbf{97.47} & 96.54 & 89.85 & \textbf{98.04} \\
Geometric mean time (s)     & \textbf{5.99} & 6.16 & 45.59 & \textbf{33.13} \\
Arithmetic mean time (s)    & \textbf{18.97} & 23.58 & 164.30 & \textbf{162.14} \\
Memory-exhaustion events    & 0 & 0 & 0 & 0 \\
\midrule
Faster, small instances     & 1 & \textbf{20} & 0 & \textbf{19} \\
Faster, medium instances    & \textbf{25} & 20 & 8 & \textbf{32} \\
Faster, large instances     & \textbf{28} & 6 & \textbf{32} & 7 \\
\bottomrule
\multicolumn{5}{p{0.95\columnwidth}}{\footnotesize Size bands are defined by plan length (PL): small $\mathrm{PL} \leq 50$, medium $50 < \mathrm{PL} \leq 150$, large $\mathrm{PL} > 150$, band sizes are $21/45/34$ instances (nominal) and $20/40/39$ (disrupted).}
\end{tabular}
\end{table}

On the disrupted set, aggregate runtimes do not differ significantly (Wilcoxon, $n{=}98$, $z{=}0.87$, $p{=}0.382$, OPTIC faster on 40 instances, POPF on 58, with one tie).
POPF is faster on 19 of the 20 smallest instances (PL $\leq 50$) and on 32 of the 40 medium instances ($50 < \mathrm{PL} \leq 150$), which its higher agile score (98.04 against 89.85) and lower geometric-mean runtime (33.1\,s against 45.6\,s) reflect. 
OPTIC is faster on 32 of the 39 largest instances (PL $> 150$). 
The planners differ markedly in robustness. 
OPTIC exhausted memory on 65 of 100 instances, always \emph{after} emitting a valid plan, so coverage and quality were unaffected, indicating a memory-unstable post-solution phase at scale.
POPF terminated cleanly on all instances.
Since these failures truncate OPTIC's search statistics, search-space analysis is restricted to POPF.

\paragraph{Operational Feasibility}
Relating planning cost to the operational timescale of the disruptions being managed, the longest single planning run across both planners and all solved instances was $795$ s ($13.3$ min) on the disrupted set and $136$ s. 
Mean planning time on disrupted instances ranges from $0.3$ s at S1 to $10.7$ min at VL3.
Expressed against the delay each plan manages, the computational overhead of replanning amounts to between 0.02\% (S1) and 2.9\% (VL2) of the average disruption-induced delay per instance. 
Recovery-plan computation is therefore small relative to the operational timescale of the disruptions themselves, which supports the suitability of the framework for disruption response on dispatching second-level response. 
This conclusion is bounded by the benchmark: the unsolved instance p200 shows that the memory budget of the reported configuration is exhausted at the extreme end of the scale range.

\subsection{Correlation Analysis}
\label{sec:correlation}
To characterize how problem scale and disruption structure drive performance, we computed Spearman rank correlations at the level of individual instances ($n = 99$ commonly solved disrupted instances) presented in Table~\ref{tab:instance_correlations}.
We use plan length as the problem-scale proxy, since it grows deterministically with the number of trains, track points, and junction controllers in the benchmark design. 
Instance-level analysis is essential here: correlations computed on scale-group aggregates are trivially close to $\pm 1$ for any quantity monotone in scale and would overstate the strength of the underlying relationships.

\begin{table*}[!t]
\centering
\caption{Spearman correlation heatmap over the $n{=}99$ solved disrupted-operation instances. 
Coefficients with $|\rho| \geq 0.53$ are significant at $p < 10^{-9}$, coefficients below $0.20$ are not significant at the $0.05$ level.}
\label{tab:instance_correlations}
\small
\setlength{\tabcolsep}{2pt}
\renewcommand{\arraystretch}{1.3}
\begin{tabular}{lcccccccc c c}
& & & & & & & & & & \\[0.3cm]
\textbf{Plan Length}
 & \cellcolor{gray!60}\textbf{1.00} & & & & & & & & \text{~~}
 & \cellcolor{gray!60}~~1.0~~ \\[0.1cm]
\textbf{Makespan}
 & \cellcolor{gray!34}\textbf{0.57} & \cellcolor{gray!60}\textbf{1.00}
 & & & & & & & & \cellcolor{gray!54}~~0.9~~ \\[0.1cm]
\textbf{Total Delay}
 & \cellcolor{gray!51}\textbf{0.85} & \cellcolor{gray!45}\textbf{0.75}
 & \cellcolor{gray!60}\textbf{1.00} & & & & & &
 & \cellcolor{gray!48}~~0.8~~ \\[0.1cm]
\textbf{Slowdown Delay}
 & \cellcolor{gray!7}\textbf{0.11} & \cellcolor{gray!33}\textbf{0.55}
 & \cellcolor{gray!32}\textbf{0.53} & \cellcolor{gray!60}\textbf{1.00}
 & & & & & & \cellcolor{gray!42}~~0.7~~ \\[0.1cm]
\textbf{Blockage Delay}
 & \cellcolor{gray!59}\textbf{0.98} & \cellcolor{gray!35}\textbf{0.58}
 & \cellcolor{gray!52}\textbf{0.87} & \cellcolor{gray!8}\textbf{0.13}
 & \cellcolor{gray!60}\textbf{1.00} & & & &
 & \cellcolor{gray!36}~~0.6~~ \\[0.1cm]
\textbf{Engine Failure Delay}
 & \cellcolor{gray!59}\textbf{0.98} & \cellcolor{gray!33}\textbf{0.55}
 & \cellcolor{gray!50}\textbf{0.84} & \cellcolor{gray!5}\textbf{0.09}
 & \cellcolor{gray!59}\textbf{0.98} & \cellcolor{gray!60}\textbf{1.00}
 & & & & \cellcolor{gray!24}~~0.4~~ \\[0.1cm]
\textbf{Time (POPF)}
 & \cellcolor{gray!59}\textbf{0.99} & \cellcolor{gray!34}\textbf{0.57}
 & \cellcolor{gray!51}\textbf{0.85} & \cellcolor{gray!7}\textbf{0.11}
 & \cellcolor{gray!58}\textbf{0.97} & \cellcolor{gray!59}\textbf{0.98}
 & \cellcolor{gray!60}\textbf{1.00} & &
 & \cellcolor{gray!12}~~0.2~~ \\[0.1cm]
\textbf{Time (OPTIC)}
 & \cellcolor{gray!57}\textbf{0.95} & \cellcolor{gray!33}\textbf{0.55}
 & \cellcolor{gray!49}\textbf{0.81} & \cellcolor{gray!7}\textbf{0.11}
 & \cellcolor{gray!56}\textbf{0.93} & \cellcolor{gray!57}\textbf{0.95}
 & \cellcolor{gray!58}\textbf{0.96} & \cellcolor{gray!60}\textbf{1.00}
 & & \cellcolor{gray!5}~~0.0~~ \\[0.3cm]
& \rotatebox{90}{\textbf{~Plan Length~}} & \rotatebox{90}{\textbf{~Makespan~}}
& \rotatebox{90}{\textbf{~Total Delay~}} & \rotatebox{90}{\textbf{~Slowdown Delay~}}
& \rotatebox{90}{\textbf{~Blockage Delay~}} & \rotatebox{90}{\textbf{~Engine Failure Delay~}}
& \rotatebox{90}{\textbf{~Time (POPF)~}}
& \rotatebox{90}{\textbf{~Time (OPTIC)~}} & & \\
\end{tabular}
\end{table*}

Three patterns emerge. 
First, computation time is driven almost entirely by problem scale ($\rho = 0.99$ with plan length for POPF, $0.95$ for OPTIC) rather than by disruption severity, indicating that replanning cost in this domain is a function of network size, not of the amount of disruption to be absorbed. 
Second, blockage and engine-repair delays correlate almost perfectly with scale ($\rho = 0.98$), consistent with the benchmark design in which the number of disruption events grows with network size. 
Third, and in contrast, slowdown delay is nearly scale-independent ($\rho = 0.11$, not significant) and instead tracks makespan ($\rho = 0.55$): the severity of slowdown-induced delay depends on which track segments are degraded and how central they are to the traffic flow, rather than on network size.
Total delay correlates strongly with both scale ($\rho = 0.85$) and makespan ($\rho = 0.75$), confirming that the delay metrics capture genuine schedule degradation rather than an artifact of instance size alone.

\section{CONCLUSIONS}
\label{sec:conclusion}
In this study, a temporal planning-based framework for dynamic route optimization and disruption management in heterogeneous multi-gauge railway networks is developed. 

Our framework formulates railway operations in PDDL 2.1 to produce timestamped action plans with explicit constraints for gauge compatibility, operational safety, and automated disruption resolution. 
An extensive evaluation on 200 benchmark instances shows that two temporal planners solve 99-100\% of the instances with identical plan quality on commodity hardware, exhibit a size-dependent runtime crossover, and generate recovery schedules whose delay decomposition and resilience characteristics scale consistently with network size and disruption intensity.
The results confirm the potential of temporal planning to generate feasible and safety-compliant operational strategies in real-time even in large scale and disruption prone railway environments.

Overall, this work establishes temporal planning as a practical and scalable foundation for intelligent railway management systems. 
In future work, we intend to extend the framework along the improvement paths identified by our analysis: domain-specific heuristics and problem decomposition to push the observed scaling boundary, and plan repair and anytime replanning under uncertainty to handle incomplete or evolving operational information. 
Additional extensions, including derailment modeling and integration with real-time data streams, will further enhance the framework's resilience and deployment readiness.



\section{Data Availability}
\label{sec:data}
The planning domain, problem instances, results, and validation outputs are available at
\url{https://github.com/PollobRay/Railway-Route-Planning}.

\bibliography{sn-bibliography}

@article{islam2024optimizing,
title = {Optimizing hybrid renewable energy based automated railway level crossing in Bangladesh: Techno-economic, emission and sensitivity analysis},
journal = {Energy Conversion and Management: X},
volume = {24},
pages = {100744},
year = {2024},
issn = {2590-1745},
doi = {https://doi.org/10.1016/j.ecmx.2024.100744},
url = {https://www.sciencedirect.com/science/article/pii/S2590174524002228},
author = {Zia {Ul Islam} and M.S. {Hossain Lipu} and Tahia F. Karim and Abu M. Fuad and M.M. Naushad Ali and ASM Shihavuddin and Ahmed {Al Mansur}},
}

@article{leutwiler2023review,
title = {A review of principles and methods to decompose large-scale railway scheduling problems},
journal = {EURO Journal on Transportation and Logistics},
volume = {12},
pages = {100107},
year = {2023},
issn = {2192-4376},
doi = {https://doi.org/10.1016/j.ejtl.2023.100107},
url = {https://www.sciencedirect.com/science/article/pii/S2192437623000043},
author = {Florin Leutwiler and Francesco Corman}
}

@techreport{br2023,
  author       = {{Bangladesh Railway}},
  title        = {Information Book 2023},
  institution  = {Bangladesh Railway},
  address      = {Dhaka},
  year         = {2023},
  url          = {https://railway.portal.gov.bd/sites/default/files/files/railway.portal.gov.bd/files/e20c67d6_51d7_47b1_905e_0cdbc1b7b529/Information%20Book%202023%20Full%20Book%20Final%20Copy.pdf},
  note         = {Accessed: March 12, 2026}
}

@article{xiu2024passenger,
title = {Passenger service-oriented timetable rescheduling for large-scale disruptions in a railway network: A heuristic-based alternating direction method of multipliers},
journal = {Omega},
volume = {125},
pages = {103040},
year = {2024},
issn = {0305-0483},
doi = {https://doi.org/10.1016/j.omega.2024.103040},
url = {https://www.sciencedirect.com/science/article/pii/S0305048324000070},
author = {Cong Xiu and Jinyi Pan and Andrea D'Ariano and Shuguang Zhan and Qiyuan Peng}
}

@article{li2025rolling,
title = {Rolling stock rescheduling in high-speed railway networks via a nested Benders decomposition approach},
journal = {Transportation Research Part C: Emerging Technologies},
volume = {171},
pages = {105001},
year = {2025},
issn = {0968-090X},
doi = {https://doi.org/10.1016/j.trc.2025.105001},
url = {https://www.sciencedirect.com/science/article/pii/S0968090X25000051},
author = {Denghui Li and Jun Zhao and Qiyuan Peng and Dian Wang and Qingwei Zhong}
}

@article{neves2024human,
title = {Human reliability and Organizational factors—How do Human Factors contribute to Signals Passed at Danger?},
journal = {Safety Science},
volume = {171},
pages = {106395},
year = {2024},
issn = {0925-7535},
doi = {https://doi.org/10.1016/j.ssci.2023.106395},
url = {https://www.sciencedirect.com/science/article/pii/S0925753523003375},
author = {Gonçalo Neves and Guilherme Ribeiro and Miguel Grilo and Virgínia Infante and António R. Andrade}
}

@article{lemos2024iterative,
author = {Lemos, Alexandre and Gouveia, Filipe and Monteiro, Pedro T. and Lynce, In\^{e}s},
title = {Iterative Train Scheduling under Disruption with Maximum Satisfiability},
year = {2024},
issue_date = {Apr 2024},
publisher = {AI Access Foundation},
address = {El Segundo, CA, USA},
volume = {79},
issn = {1076-9757},
url = {https://doi.org/10.1613/jair.1.14924},
doi = {10.1613/jair.1.14924},
journal = {J. Artif. Int. Res.},
month = apr,
numpages = {44}
}

@article{chai2024branch,
title = {A branch-and-cut algorithm for scheduling train platoons in urban rail networks},
journal = {Transportation Research Part B: Methodological},
volume = {181},
pages = {102891},
year = {2024},
issn = {0191-2615},
doi = {https://doi.org/10.1016/j.trb.2024.102891},
url = {https://www.sciencedirect.com/science/article/pii/S0191261524000158},
author = {Simin Chai and Jiateng Yin and Andrea D’Ariano and Ronghui Liu and Lixing Yang and Tao Tang}
}

@article{ji2024optimization,
title = {Optimization of train schedule with uncertain maintenance plans in high-speed railways: A stochastic programming approach},
journal = {Omega},
volume = {124},
pages = {102999},
year = {2024},
issn = {0305-0483},
doi = {https://doi.org/10.1016/j.omega.2023.102999},
url = {https://www.sciencedirect.com/science/article/pii/S0305048323001639},
author = {Hangyu Ji and Rui Wang and Chuntian Zhang and Jiateng Yin and Lin Ma and Lixing Yang}
}

@article{zhang2022train,
author = {Zhang, Hai and Ni, Shaoquan},
title = {Train Scheduling Optimization for an Urban Rail Transit Line: A Simulated-Annealing Algorithm Using a Large Neighborhood Search Metaheuristic},
journal = {Journal of Advanced Transportation},
volume = {2022},
number = {1},
pages = {9604362},
doi = {https://doi.org/10.1155/2022/9604362},
url = {https://onlinelibrary.wiley.com/doi/abs/10.1155/2022/9604362},
eprint = {https://onlinelibrary.wiley.com/doi/pdf/10.1155/2022/9604362},
year = {2022}
}

@article{zhuo2024demand,
title = {Demand-driven integrated train timetabling and rolling stock scheduling on urban rail transit line},
journal = {Transportmetrica A Transport Science},
volume = {20},
number = {3},
year = {2024},
issn = {2324-9935},
doi = {https://doi.org/10.1080/23249935.2023.2181024},
url = {https://www.sciencedirect.com/science/article/pii/S2324993523000386},
author = {Siyu Zhuo and Jianrui Miao and Lingyun Meng and Liya Yang and Pan Shang}
}

@article{acuna2011sapi,
title = {SAPI: Statistical Analysis of Propagation of Incidents. A new approach for rescheduling trains after disruptions},
journal = {European Journal of Operational Research},
volume = {215},
number = {1},
pages = {227-243},
year = {2011},
issn = {0377-2217},
doi = {https://doi.org/10.1016/j.ejor.2011.05.047},
url = {https://www.sciencedirect.com/science/article/pii/S0377221711004954},
author = {Rodrigo Acuna-Agost and Philippe Michelon and Dominique Feillet and Serigne Gueye}
}

@article{veelenturf2016railway,
author = {Veelenturf, Lucas P. and Kidd, Martin P. and Cacchiani, Valentina and Kroon, Leo G. and Toth, Paolo},
title = {A Railway Timetable Rescheduling Approach for Handling Large-Scale Disruptions},
journal = {Transportation Science},
volume = {50},
number = {3},
pages = {841-862},
year = {2016},
doi = {10.1287/trsc.2015.0618},
URL = {https://doi.org/10.1287/trsc.2015.0618},
eprint = {https://doi.org/10.1287/trsc.2015.0618}
}

@article{fischetti2017using,
title = {Using a general-purpose Mixed-Integer Linear Programming solver for the practical solution of real-time train rescheduling},
journal = {European Journal of Operational Research},
volume = {263},
number = {1},
pages = {258-264},
year = {2017},
issn = {0377-2217},
doi = {https://doi.org/10.1016/j.ejor.2017.04.057},
url = {https://www.sciencedirect.com/science/article/pii/S0377221717304046},
author = {Matteo Fischetti and Michele Monaci}
}

@article{benton2012temporal, title={Temporal Planning with Preferences and Time-Dependent Continuous Costs}, volume={22}, url={https://ojs.aaai.org/index.php/ICAPS/article/view/13509}, DOI={10.1609/icaps.v22i1.13509}, number={1}, journal={Proceedings of the International Conference on Automated Planning and Scheduling}, author={Benton, J. and Coles, Amanda and Coles, Andrew}, year={2012}, month={May}, pages={2-10} }

@incollection{ghallab2004automated,
title = {Chapter 1 - Introduction and Overview},
editor = {Malik Ghallab and Dana Nau and Paolo Traverso},
booktitle = {Automated Planning},
publisher = {Morgan Kaufmann},
address = {Burlington},
pages = {1-16},
year = {2004},
series = {The Morgan Kaufmann Series in Artificial Intelligence},
isbn = {978-1-55860-856-6},
doi = {https://doi.org/10.1016/B978-155860856-6/50004-1},
url = {https://www.sciencedirect.com/science/article/pii/B9781558608566500041},
author = {Malik Ghallab and Dana Nau and Paolo Traverso},
}

@article{cenamor2019predictability,
author = {Cenamor, Isabel and Vallati, Mauro and Chrpa, Lukáš},
title = {On the predictability of domain-independent temporal planners},
journal = {Computational Intelligence},
volume = {35},
number = {4},
pages = {745-773},
doi = {https://doi.org/10.1111/coin.12211},
url = {https://onlinelibrary.wiley.com/doi/abs/10.1111/coin.12211},
eprint = {https://onlinelibrary.wiley.com/doi/pdf/10.1111/coin.12211},
year = {2019}
}

@Inbook{haslum2019introduction,
author="Haslum, Patrik
and Lipovetzky, Nir
and Magazzeni, Daniele
and Muise, Christian",
title="Temporal Planning",
bookTitle="An Introduction to the Planning Domain Definition Language",
year="2019",
publisher="Springer International Publishing",
address="Cham",
pages="103--122",
isbn="978-3-031-01584-7",
doi="10.1007/978-3-031-01584-7\_5",
url="https://doi.org/10.1007/978-3-031-01584-7\_5"
}

@article{fox2003pddl2,
author = {Fox, Maria and Long, Derek},
title = {PDDL2.1: an extension to PDDL for expressing temporal planning domains},
year = {2003},
issue_date = {December 2003},
publisher = {AI Access Foundation},
address = {El Segundo, CA, USA},
volume = {20},
number = {1},
issn = {1076-9757},
journal = {J. Artif. Int. Res.},
month = dec,
pages = {61–124},
numpages = {64}
}

@article{nitisiri2019parallel,
title = {A parallel multi-objective genetic algorithm with learning based mutation for railway scheduling},
journal = {Computers \& Industrial Engineering},
volume = {130},
pages = {381-394},
year = {2019},
issn = {0360-8352},
doi = {https://doi.org/10.1016/j.cie.2019.02.035},
url = {https://www.sciencedirect.com/science/article/pii/S0360835219301214},
author = {Krisanarach Nitisiri and Mitsuo Gen and Hayato Ohwada}
}

@article{cardellini2021station, 
title={In-Station Train Dispatching: A PDDL+ Planning Approach}, 
volume={31}, 
url={https://ojs.aaai.org/index.php/ICAPS/article/view/15991}, 
DOI={10.1609/icaps.v31i1.15991},  
number={1}, 
journal={Proceedings of the International Conference on Automated Planning and Scheduling}, 
author={Cardellini, Matteo and Maratea, Marco and Vallati, Mauro and Boleto, Gianluca and Oneto, Luca}, 
year={2021}, month={May}, pages={450-458} 
}

@inproceedings{louadah2021translating,
title = "Translating ontological knowledge to PDDL to do Planning in Train Depot Management Operations",
author = "Hassna Louadah and Emmanuel Papadakis and Lee McCluskey and Gareth Tucker and Peter Hughes and Adam Bevan",
year = "2021",
month = dec,
day = "20",
language = "English",
booktitle = "Proceedings of PlanSIG 2021",
publisher = "AAAI press",
address = "United States",
note = "36th Workshop of the UK Planning and Scheduling Special Interest Group, PlanSIG 2021 ; Conference date: 20-12-2021 Through 20-12-2021",
url = "https://plansig2021.wordpress.com/",
}

@article{coles2011popf2, 
title={Forward-Chaining Partial-Order Planning}, 
volume={20}, 
url={https://ojs.aaai.org/index.php/ICAPS/article/view/13403}, 
DOI={10.1609/icaps.v20i1.13403},  
number={1}, 
journal={Proceedings of the International Conference on Automated Planning and Scheduling}, 
author={Coles, Amanda and Coles, Andrew and Fox, Maria and Long, Derek}, year={2010}, month={May}, pages={42-49} }

@book{bryan2007rail,
  author    = "{{Transportation Research Board and National Academies of Sciences, Engineering, and Medicine}}",
  title     = {Rail Freight Solutions to Roadway Congestion--Final Report and Guidebook},
  doi       = {10.17226/14098},
  url       = {https://nap.nationalacademies.org/catalog/14098/rail-freight-solutions-to-roadway-congestion-final-report-and-guidebook},
  year      = {2007},
  publisher = {The National Academies Press},
  address   = {Washington, DC}
}

@ARTICLE{kang2024critical,
  author={Kang, Liujiang and Buhigiro, Nsabimana and Sun, Huijun and Wu, Jianjun},
  journal={IEEE Transactions on Intelligent Transportation Systems}, 
  title={A Critical Review of Subway Train Timetabling and Rescheduling Problems}, 
  year={2024},
  volume={25},
  number={10},
  pages={12930-12942},
  doi={10.1109/TITS.2024.3386728}
}

@Article{wang2022train,
AUTHOR = {Wang, Yidong and Song, Rui and He, Shiwei and Song, Zilong},
TITLE = {Train Routing and Track Allocation Optimization Model of Multi-Station High-Speed Railway Hub},
JOURNAL = {Sustainability},
VOLUME = {14},
YEAR = {2022},
NUMBER = {12},
ARTICLE-NUMBER = {7292},
URL = {https://www.mdpi.com/2071-1050/14/12/7292},
ISSN = {2071-1050},
DOI = {10.3390/su14127292}
}

@article{sanchis2021experimental,
title = {Experimental and numerical investigations of dual gauge railway track behaviour},
journal = {Construction and Building Materials},
volume = {299},
pages = {123943},
year = {2021},
issn = {0950-0618},
doi = {https://doi.org/10.1016/j.conbuildmat.2021.123943},
url = {https://www.sciencedirect.com/science/article/pii/S0950061821017037},
author = {Ignacio {Villalba Sanchis} and Ricardo {Insa Franco} and Pablo {Martínez Fernández} and Pablo {Salvador Zuriaga}}
}

@inproceedings{gouveia2025maxsat,
author = {Gouveia, Filipe and Albino, Lu\'{\i}s and Saldanha, Ricardo L.},
title = {A MaxSAT Approach for the Train Timetabling Problem with Route Choice and Other Features},
year = {2025},
isbn = {978-3-032-05178-3},
publisher = {Springer-Verlag},
address = {Berlin, Heidelberg},
url = {https://doi.org/10.1007/978-3-032-05179-0\_14},
doi = {10.1007/978-3-032-05179-0\_14},
booktitle = {Progress in Artificial Intelligence: 24th EPIA Conference on Artificial Intelligence, EPIA 2025, Faro, Portugal, October 1--3, 2025, Proceedings, Part II},
pages = {179--191},
numpages = {13},
location = {Faro, Portugal}
}

@INPROCEEDINGS{coviello2023integrated,
  author={Coviello, Nicola and Medeossi, Giorgio and Nygreen, Thomas and Pellegrini, Paola and Rodriguez, Joaquin},
  booktitle={2023 IEEE 26th International Conference on Intelligent Transportation Systems (ITSC)}, 
  title={Integrated Ant Colony Optimization and Mixed Integer Linear Programming for Multi-objective Railway Timetabling}, 
  year={2023},
  volume={},
  number={},
  pages={4973-4979},
  doi={10.1109/ITSC57777.2023.10421843}
  }

@Article{su17188336,
AUTHOR = {Grigonis, Vytautas and Kaušylas, Mantas and Palevičius, Vytautas},
TITLE = {Advancing Sustainable Interoperability Between Standard and Broad-Gauge Railway Systems},
JOURNAL = {Sustainability},
VOLUME = {17},
YEAR = {2025},
NUMBER = {18},
ARTICLE-NUMBER = {8336},
URL = {https://www.mdpi.com/2071-1050/17/18/8336},
ISSN = {2071-1050},
DOI = {10.3390/su17188336}
}

@inproceedings{VAL_1374201,
author = {Howey, Richard and Long, Derek and Fox, Maria},
title = {VAL: Automatic Plan Validation, Continuous Effects and Mixed Initiative Planning Using PDDL},
year = {2004},
isbn = {076952236X},
publisher = {IEEE Computer Society},
address = {USA},
url = {https://doi.org/10.1109/ICTAI.2004.120},
doi = {10.1109/ICTAI.2004.120},
booktitle = {Proceedings of the 16th IEEE International Conference on Tools with Artificial Intelligence},
pages = {294–301},
numpages = {8},
series = {ICTAI '04}
}

@article{yan2019multi,
title = {Multi-objective periodic railway timetabling on dense heterogeneous railway corridors},
journal = {Transportation Research Part B: Methodological},
volume = {125},
pages = {52-75},
year = {2019},
issn = {0191-2615},
doi = {https://doi.org/10.1016/j.trb.2019.05.002},
url = {https://www.sciencedirect.com/science/article/pii/S0191261518308488},
author = {Fei Yan and Nikola Bešinović and Rob M.P. Goverde}
}

@article{NYPORKO2025102415,
title = {Modelling and solving industrial production tasks as planning-scheduling tasks},
journal = {Data \& Knowledge Engineering},
volume = {157},
pages = {102415},
year = {2025},
issn = {0169-023X},
doi = {https://doi.org/10.1016/j.datak.2025.102415},
url = {https://www.sciencedirect.com/science/article/pii/S0169023X25000102},
author = {Andrii Nyporko and Lukáš Chrpa}
}

@inproceedings{Yang2022,
  title={An AI Planning Approach to Emergency Material Scheduling Using Numerical PDDL},
  author={Liping Yang and Ruishi Liang},
  year={2022},
  booktitle={Proceedings of the 2022 International Conference on Artificial Intelligence, Internet and Digital Economy (ICAID 2022)},
  pages={47-54},
  issn={2589-4919},
  isbn={978-94-6463-010-7},
  url={https://doi.org/10.2991/978-94-6463-010-7\_7},
  doi={10.2991/978-94-6463-010-7\_7},
  publisher={Atlantis Press}
}

@INPROCEEDINGS{6984461,
  author={Caff, Sebastiano Concetto Marco and Di Mauro, Francesco and Scala, Enrico},
  booktitle={2014 IEEE 26th International Conference on Tools with Artificial Intelligence}, 
  title={A Numeric PDDL Based Approach for Temporally Constrained Journey Problems}, 
  year={2014},
  volume={},
  number={},
  pages={99-106},
  doi={10.1109/ICTAI.2014.25}
}

@ARTICLE{7160720,
  author={Fang, Wei and Yang, Shengxiang and Yao, Xin},
  journal={IEEE Transactions on Intelligent Transportation Systems}, 
  title={A Survey on Problem Models and Solution Approaches to Rescheduling in Railway Networks}, 
  year={2015},
  volume={16},
  number={6},
  pages={2997-3016},
  doi={10.1109/TITS.2015.2446985}
}

@article{ALBRECHT2013703,
title = {Rescheduling rail networks with maintenance disruptions using Problem Space Search},
journal = {Computers \& Operations Research},
volume = {40},
number = {3},
pages = {703-712},
year = {2013},
note = {Transport Scheduling},
issn = {0305-0548},
doi = {https://doi.org/10.1016/j.cor.2010.09.001},
url = {https://www.sciencedirect.com/science/article/pii/S0305054810001905},
author = {A.R. Albrecht and D.M. Panton and D.H. Lee}
}

@article{DUNDAR20131,
title = {Train re-scheduling with genetic algorithms and artificial neural networks for single-track railways},
journal = {Transportation Research Part C: Emerging Technologies},
volume = {27},
pages = {1-15},
year = {2013},
note = {Selected papers from the Seventh Triennial Symposium on Transportation Analysis (TRISTAN VII)},
issn = {0968-090X},
doi = {https://doi.org/10.1016/j.trc.2012.11.001},
url = {https://www.sciencedirect.com/science/article/pii/S0968090X12001349},
author = {Selim Dündar and İsmail Şahin}
}

@article{20229604362,
author = {Zhang, Hai and Ni, Shaoquan},
title = {Train Scheduling Optimization for an Urban Rail Transit Line: A Simulated-Annealing Algorithm Using a Large Neighborhood Search Metaheuristic},
journal = {Journal of Advanced Transportation},
volume = {2022},
number = {1},
pages = {9604362},
doi = {https://doi.org/10.1155/2022/9604362},
url = {https://onlinelibrary.wiley.com/doi/abs/10.1155/2022/9604362},
eprint = {https://onlinelibrary.wiley.com/doi/pdf/10.1155/2022/9604362},
year = {2022}
}

@article{Narayanaswami2011,
  author  = {Sundaravalli Narayanaswami and Narayan Rangaraj},
  title   = {Scheduling and Rescheduling of Railway Operations: A Review and Expository Analysis},
  journal = {Technology Operation Management},
  year    = {2011},
  volume  = {2},
  number  = {2},
  pages   = {102--122},
  doi     = {10.1007/s13727-012-0006-x},
  url     = {https://doi.org/10.1007/s13727-012-0006-x},
  issn    = {2249-2364}
}

\end{document}